\documentclass[10pt,twocolumn,letterpaper]{article}

\usepackage{iccv}
\usepackage{times}
\usepackage{epsfig}
\usepackage{graphicx}
\usepackage{amsmath}
\usepackage{amssymb}

\usepackage[breaklinks=true,bookmarks=false]{hyperref}

\iccvfinalcopy 

\def\iccvPaperID{****} 

\ificcvfinal\fi

\usepackage{siunitx}    
\usepackage{array,multirow}
\usepackage{adjustbox}
\usepackage{colortbl}
\usepackage{xcolor}
\usepackage{arydshln}
\usepackage{caption}
\usepackage{acro}
\DeclareAcronym{uq}{
    short=UQ,
    long=Uncertainty Quantification
}
\DeclareAcronym{bnn}{
    short=BNN,
    long=Bayesian Neural Networks
}
\DeclareAcronym{mcdropout}{
    short=MC-Dropout,
    long=Monte Carlo Dropout
}
\DeclareAcronym{vae}{
    short=VAE,
    long=Variational Autoencoders
}
\DeclareAcronym{de}{
    short=DE,
    long=Deep Ensembles
}
\DeclareAcronym{ud}{
    short=UD,
    long=Uncertainty Distillation
}
\DeclareAcronym{kd}{
    short=KD,
    long=Knowledge Distillation
}
\DeclareAcronym{dudes}{
    short=DUDES,
    long='\textbf{D}eep \textbf{U}ncertainty \textbf{D}istillation using \textbf{E}nsembles for \textbf{S}egmentation'
}
\DeclareAcronym{sgd}{
    short=SGD,
    long=Stochastic Gradient Descent
}
\DeclareAcronym{dnn}{
    short=DNN,
    long=Deep Neural Network
}
\DeclareAcronym{rmsle}{
    short=RMSLE,
    long=root mean squared logarithmic error,
}

\begin{document}

\title{DUDES: Deep Uncertainty Distillation using Ensembles\\ for Semantic Segmentation}

\author{Steven Landgraf \hspace{2mm} Kira Wursthorn \hspace{2mm} Markus Hillemann \hspace{2mm} Markus Ulrich\\
Karlsruhe Institute of Technology (KIT)\\
{\tt\small (steven.landgraf, kira.wursthorn, markus.hillemann, markus.ulrich)@kit.edu}
}

\maketitle
\ificcvfinal\thispagestyle{empty}\fi

\begin{abstract}
Deep neural networks lack interpretability and tend to be overconfident, which poses a serious problem in safety-critical applications like autonomous driving, medical imaging, or machine vision tasks with high demands on reliability. Quantifying the predictive uncertainty is a promising endeavour to open up the use of deep neural networks for such applications. Unfortunately, current available methods are computationally expensive. In this work, we present a novel approach for efficient and reliable uncertainty estimation which we call \textbf{D}eep \textbf{U}ncertainty \textbf{D}istillation using \textbf{E}nsembles for \textbf{S}egmentation (DUDES). DUDES applies student-teacher distillation with a Deep Ensemble to accurately approximate predictive uncertainties with a single forward pass while maintaining simplicity and adaptability. Experimentally, DUDES accurately captures predictive uncertainties without sacrificing performance on the segmentation task and indicates impressive capabilities of identifying wrongly classified pixels and out-of-domain samples on the Cityscapes dataset. With DUDES, we manage to simultaneously simplify and outperform previous work on Deep Ensemble-based Uncertainty Distillation.
\end{abstract}

\section{Introduction}
\label{sec:intro}

\begin{figure}[ht!]
\centering
\includegraphics[width=\linewidth]{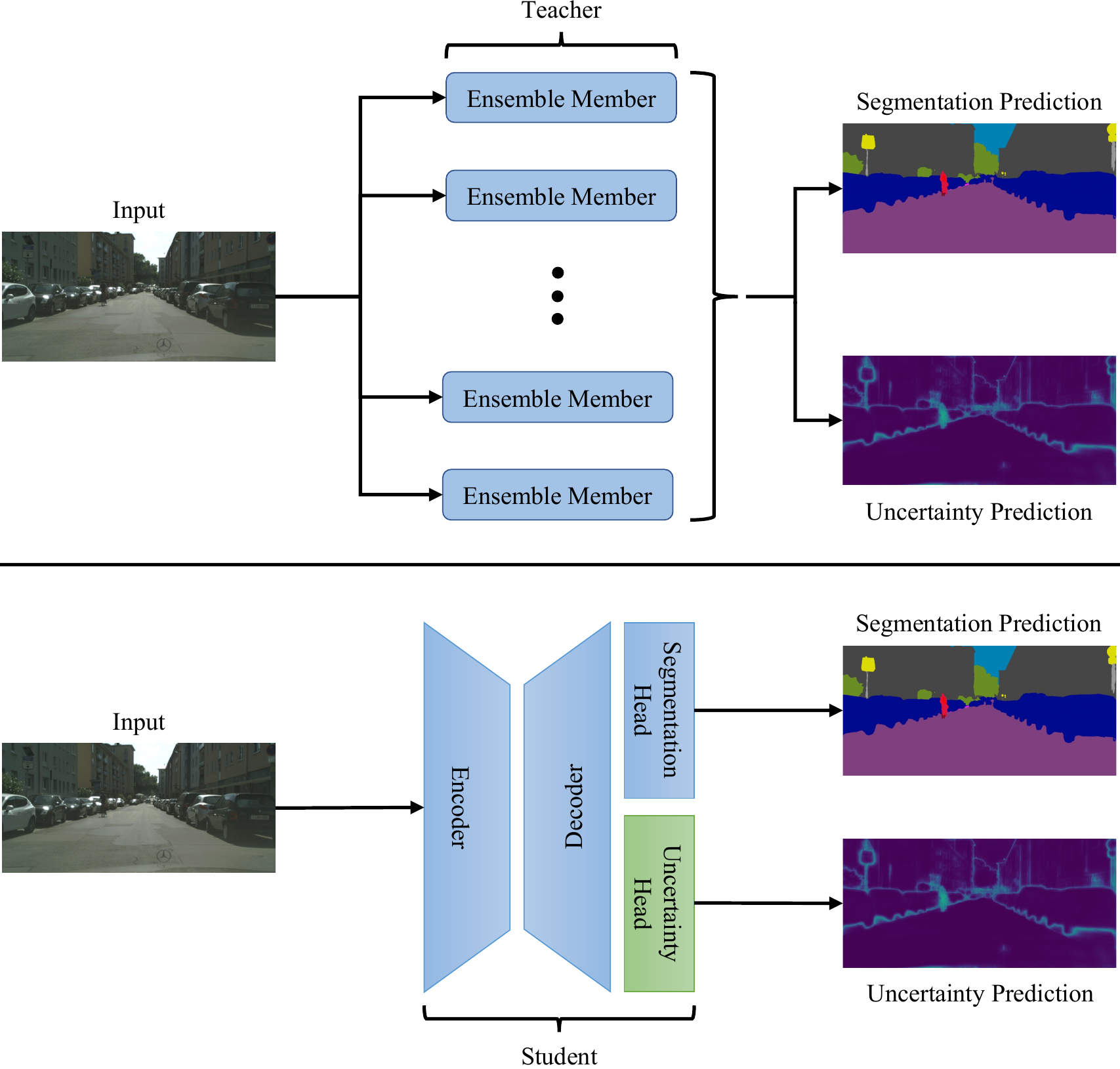}
\caption{\ac{dudes} applies student-teacher distillation with a Deep Ensemble (DE) to accurately approximate predictive uncertainties with a single forward pass while maintaining simplicity and adaptability.}
\label{fig: Student}
\end{figure}

Semantic segmentation is a computer vision task that aims to assign a class label to each pixel in a given image, with the goal of understanding the semantic content of the image. Hence, it can be viewed as a pixel-wise classification task. Recently, methods based on deep neural networks have become the most popular and successful approach to solve this problem \cite{minaee2020ImageSegmentation}. Despite their unrivaled performance on established benchmark datasets like Cityscapes \cite{cordts2016CityscapesDataset} or PASCAL VOC \cite{everingham2015PascalVisual}, neural networks lack interpretability \cite{gawlikowski2022SurveyUncertainty}, are unable to distinguish between in-domain and out-of-domain samples \cite{lee2018TrainingConfidencecalibrated}, and tend to be overconfident \cite{guo2017CalibrationModerna}. These shortcomings are especially severe for safety-critical applications like autonomous driving \cite{mcallister2017ConcreteProblems} and the analysis of medical imaging \cite{leibig2017LeveragingUncertainty} or computer vision tasks that have high demands on reliability like industrial inspection \cite{steger2018MachineVision} and automation \cite{ulrich_2021}.

Quantifying the predictive uncertainty is a promising endeavour to make such applications safer and more reliable, e.g., by preemptively making risk-averse predictions or by providing feedback to a human operator when predictions are uncertain. Some of the most relevant methods include Bayesian Neural Networks \cite{mackay1992PracticalBayesian}, Monte Carlo Dropout \cite{gal2016DropoutBayesian}, and \ac{de} \cite{lakshminarayanan2017SimpleScalable}. Unfortunately, all of these methods require computationally expensive estimation of a distribution of outputs by sampling from a stochastic process. Recently, the concept of \ac{kd} has been introduced as a potential solution \cite{shen2020RealTimeUncertaintya,besnier2021LearningUncertainty,holder2021EfficientUncertainty,simpson2022LearningStructured}. Knowledge distillation is a technique for transferring the knowledge embodied in a complex model, referred to as the teacher, to a smaller model, referred to as the student. By incorporating the knowledge learned by a more complex model, the student's performance can be enhanced \cite{hinton2015DistillingKnowledgea, romero2015FitNetsHints, malinin2019EnsembleDistributiona}. 

In this work, we present a novel approach for efficient and reliable uncertainty quantification, which we call \ac{dudes} as shown in Figure \ref{fig: Student}. \ac{dudes} applies student-teacher distillation with a \ac{de} to accurately approximate predictive uncertainties while maintaining simplicity and adaptability. In comparison to the \ac{de}, only a single forward pass is required to obtain predictive uncertainties, which massively reduces the inference time and eliminates the computational overhead. \ac{dudes} simultaneously simplifies and outperforms previous work on \ac{de}-based uncertainty distillation, which we experimentally evaluate on the Cityscapes dataset. Additionally, it is worth noting that, the \ac{de} can in principle be substituted by any other \ac{uq} method.

After a brief overview of the related works on \ac{uq} and \ac{kd} in Section \ref{sec:related work}, the methodology of \ac{dudes} is described in Section \ref{sec: methodology}. In Section \ref{sec: results}, we demonstrate the ability of \ac{dudes} through quantitative and qualitative analysis of the predictive uncertainties, and the potential to identify wrongly classified pixels and out-of-domain samples on the Cityscapes dataset. Section \ref{sec: discussion} discusses the experimental results and their potential impact on future research, while Section \ref{sec: conclusion} concludes the paper.  

\section{Related Work}
\label{sec:related work}

\begin{figure*}[ht!]
\centering
\includegraphics[width=\textwidth]{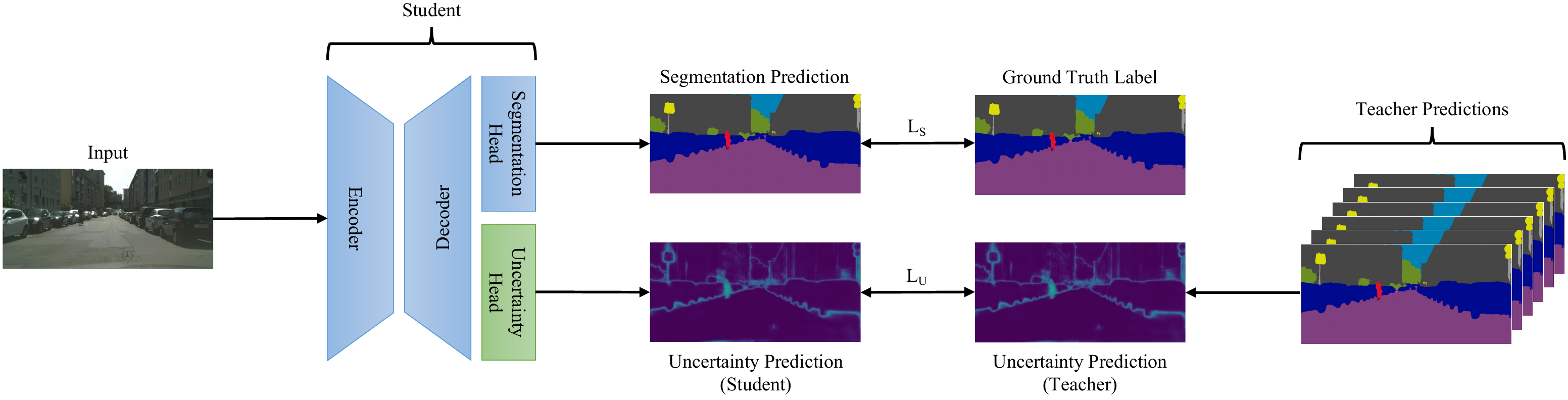}
\caption{A schematic overview of the training process of \ac{dudes}. \ac{dudes} is an easy-to-adapt framework for efficiently estimating predictive uncertainty through student-teacher distillation. The student model simultaneously outputs a segmentation prediction alongside a corresponding uncertainty prediction. Training the student involves a regular segmentation loss with the ground truth labels and an additional uncertainty loss. As ground truth uncertainties, we compute the predictive uncertainty of a \ac{de}, thereby acting as the teacher.}
\label{fig: DUDES}
\end{figure*}

In this section, we briefly summarize foundational work to \ac{dudes}. The two main methodological components of our approach are \ac{uq} (cf. Section \ref{subsec: uq}) and \ac{kd} (cf. Section \ref{subsec: kd}). 

\subsection{Uncertainty Quantification}\label{subsec: uq}
Deep neural networks consist of a large number of model parameters and often include non-linearities, which generally makes the exact posterior probability distribution of a network's output prediction intractable \cite{blundell2015BayesByBackprop,loquercio2020UncertaintyFrameworkDronet}. 
This leads to approximate \ac{uq} approaches including softmax probability, Bayesian techniques like \ac{bnn}s \cite{mackay1992PracticalBayesian} and Monte Carlo Dropout \cite{gal2016DropoutBayesian} as well as \ac{de}s \cite{lakshminarayanan2017SimpleScalable}.

While the softmax predictions are easy to implement, the predicted probabilities tend to be overconfident and need to be calibrated \cite{guo2017CalibrationModerna}. Additionally, they are often erroneously interpreted as model confidence \cite{gal2016DropoutBayesian}.
A mathematically sound approach based on Bayesian inference is provided by \ac{bnn}s where a deterministic network is transformed into a stochastic one. This is done by either placing probability distributions over the activations or the weights \cite{jospin2020HandsOnBaysianNetworks}.
For example, Bayes by Backprop \cite{blundell2015BayesByBackprop} uses variational inference to learn approximate distributions over the weights. At test time, weights are sampled from the learned distributions, resulting in an ensemble of models that is used to sample from the posterior distribution over the predictions.
To overcome the high computational cost of \ac{bnn}s, Gal and Ghahramani \cite{gal2016DropoutBayesian} propose Monte Carlo Dropout as an approximation of a stochastic gaussian process using a common regularization method. While dropout regularization \cite{srivastava2014Dropout} is usually only used during training, Monte Carlo Dropout applies this technique to sample from the posterior distribution of the predictions at test time. 
Since Monte Carlo Dropout only captures the uncertainty inherent to the model, the method is combined with learned uncertainty predictions and assumed density filtering \cite{gast2018ADF} by Kendall and Gal and Loquercio \etal\cite{kendall2017CVUncertainties,loquercio2020UncertaintyFrameworkDronet} to obtain the total uncertainty in the predictions respectively.

The uncertainties produced by Monte Carlo Dropout are not calibrated \cite{gal2016DropoutBayesian}, which is a major drawback that is overcome by \ac{de}s \cite{lakshminarayanan2017SimpleScalable} where an ensemble of trained models produces samples of predictions at test time. Due to randomness introduced by random weight initialization or different data augmentations across the ensemble members \cite{fort2020DeepEnsembles}, \ac{de}s are well-calibrated \cite{lakshminarayanan2017SimpleScalable} and outperform \ac{uq} approaches like softmax probability, Monte Carlo Dropout, and Bayes by Backprop, as is shown by Ovadia \etal\cite{ovadia2019DatasetShift}. The latter also show that \ac{de}s seem to be more robust against dataset shift, which was also observed by Wursthorn \etal\cite{wursthorn2022}.

\subsection{Knowledge Distillation}\label{subsec: kd}
\ac{kd} is a technique for transferring the knowledge embodied in a complex model, referred to as the teacher, to a smaller model, referred to as the student. The teacher can be a model with more parameters or even a \ac{de}. The student is trained to imitate the predictions of the teacher on a given dataset, with the goal of minimizing the difference of the student's outputs and the teacher's outputs. By incorporating the knowledge learned by a more complex model, the student's performance can be enhanced. Usually, this results in a more compact student model that achieves similar performance compared to the teacher model. The most formative works on \ac{kd} were published by Hinton \etal\cite{hinton2015DistillingKnowledgea}, Romero \etal\cite{romero2015FitNetsHints}, and Malinin \etal\cite{malinin2019EnsembleDistributiona}.

Recently, the concept of \ac{kd} has attracted increasing interest in the context of efficient \ac{uq} to enable real-time uncertainty estimation \cite{shen2020RealTimeUncertaintya,besnier2021LearningUncertainty,holder2021EfficientUncertainty,simpson2022LearningStructured}. For instance, Shen \etal\cite{shen2020RealTimeUncertaintya} have used student-teacher distillation for real-time \ac{uq} based on Monte Carlo Dropout \cite{srivastava2014Dropout}. Even more related to our method is the work published by Holder and Shafique\cite{holder2021EfficientUncertainty} on \ac{de}-based student-teacher distillation for efficient \ac{uq} and out-of-domain detection. Their approach requires custom segmentation and uncertainty head architectures, two additional losses, and they introduce three new hyperparameters for the distillation process. This makes their proposed method difficult to implement and to adapt to new applications. Aside from that, their student struggles to compete with the teacher's segmentation performance and does not accurately approximate the class-wise uncertainties in some cases. In contrast, \ac{dudes} does not need custom segmentation and uncertainty head architectures, only introduces a single uncertainty loss without hyperparameters, outperforms the teacher's segmentation results, and accurately captures the teacher's uncertainties. Thereby, \ac{dudes} provides a significant improvement to all of the shortcomings of the method proposed by Holder and Shafique \cite{holder2021EfficientUncertainty}. 

\section{Methodology}\label{sec: methodology}

In the following, we provide an overview of \ac{dudes}, explain the methodology behind our uncertainty distillation approach, and lay out the implementation details. Our goal is to present an easy-to-adapt framework for efficient and reliable uncertainty estimation that utilizes student-teacher distillation.

\subsection{Overview}

\ac{dudes} is a framework for efficient \ac{uq} through student-teacher distillation. The overall goal is to train a student model that can simultaneously output a segmentation prediction and a corresponding predictive uncertainty as shown by Figure \ref{fig: Student}. We propose a two-step framework:
\begin{enumerate}
    \itemsep0em     
    \item Training the teacher with the ground truth labels.
    \item Training the student with the ground truth labels and the teacher's uncertainty predictions.
\end{enumerate}
As shown in Figure \ref{fig: DUDES}, the training of the student model consists of two loss components. The first component measures the distance between the student's segmentation prediction and the ground truth labels, while the second component measures the distance between the student's uncertainty prediction and the output of one of the \ac{uq} methods described in Section \ref{subsec: uq}. As mentioned before, we choose a \ac{de} as the teacher for the concrete implementation of \ac{dudes}. \ac{de}s are simple to implement, easily parallelizable, and require little tuning. They are well-calibrated, more robust against dataset shift, and outperform other \ac{uq} methods like Monte Carlo Dropout or Bayesian Neural Networks \cite{lakshminarayanan2017SimpleScalable, fort2020DeepEnsembles, ovadia2019DatasetShift, wursthorn2022}. However, it is worth noting that the \ac{de} can in principle be substituted by any other \ac{uq} method.

\textbf{Teacher.}
To provide meaningful uncertainties, we use a \ac{de} as the teacher, which consists of ten baseline models that are not pre-trained, thus following prior work on \ac{de}-based uncertainty quantification \cite{lakshminarayanan2017SimpleScalable, fort2020DeepEnsembles}. By randomly initializing all the parameters before training, we aim to capture different aspects of the input data distribution for each ensemble member, boosting the teacher's overall performance, robustness, and uncertainty quantification capabilities. During inference, each ensemble member produces slightly different predictions, enabling the calculation of a mean segmentation prediction and an uncertainty prediction. 

\textbf{Student.}
As our student not only has to output a segmentation prediction, but also a corresponding predictive uncertainty, we add a second head to the baseline model's decoder. We propose to use an additional uncertainty head that is identical to the regular segmentation head of the baseline model, except for the output layer. For the segmentation head, we use a softmax activation to obtain class-wise probability distributions. Whereas for the uncertainty head, we use a sigmoid activation that bounds outputs between 0 and 1. Our uncertainty head only needs one output channel instead of the number of classes, as needed by the segmentation head. Since this is a key modification to improve upon previous work by Holder and Shafique \cite{holder2021EfficientUncertainty}, we will discuss this simpflication in detail in Section \ref{sec: discussion}. In contrast to the randomly initialized ensemble members, the student's parameters are initialized with ImageNet pre-training \cite{deng2009ImageNetLargescale} to improve convergence speed.

\subsection{Uncertainty Distillation}
To efficiently estimate the predictive uncertainty of the \ac{de} with a single student model, we utilize student-teacher distillation to train our student to behave like the teacher with a combination of two losses. 

\textbf{Segmentation Loss.}
The main objective function that is being minimized for the segmentation task is the well-known categorical cross-entropy loss:
\begin{equation}\label{eq:l_seg}
L_{S} = -\sum_{i=1}^{C} y_{i} \log(p_{i}),
\end{equation}
where $L_{S}$ is the segmentation loss for a single image, $C$ is the number of classes, $y_{i}$ is the ground truth label for the $i$-th class, and $p_{i}$ is the predicted probability for the $i$-th class. The categorical cross-entropy loss measures the dissimilarity between the ground truth probability distribution and the predicted probability distribution. By minimizing this loss during training, the model is encouraged to produce pixel-wise class predictions that are as close as possible to the ground truth classes.

\textbf{Uncertainty Loss.}
For distilling the predictive uncertainties of our teacher into the student, we introduce an additional uncertainty loss, which is formulated as the \ac{rmsle}:
\begin{equation}\label{eq:l_unc}
L_{U} = \sqrt{\frac{1}{N} \sum_{i=1}^{N} (\log(z_i + 1) - \log(q_i + 1))^2},
\end{equation}
where $L_{U}$ is the uncertainty loss for a single image, $N$ is the number of pixels in the image, $z_{i}$ is the teacher's predictive uncertainty for the $i$-th pixel as ground truth, and $q_i$ is the corresponding student's uncertainty prediction. The teacher's predictive uncertainty $z_{i}$ represents the standard deviation of the softmax probabilities of the predicted class in the segmentation map. By minimizing the \ac{rmsle} during training, the student is encouraged to produce uncertainty estimates that are as close as possible to the teacher's uncertainties. Since most of the uncertainties are close to zero, the natural logarithm provides special attention to the pixels where uncertainties are higher. 

\textbf{Total Loss.}
The total loss for training our student model is the sum of the segmentation loss described in Equation \ref{eq:l_seg} and the uncertainty loss expressed in Equation \ref{eq:l_unc}:
\begin{equation}\label{eq:l}
L = L_{S} + L_{U}.
\end{equation}

\subsection{Implementation Details}
\textbf{Baseline.} 
For our baseline model, we use a DeepLabv3+ \cite{chen2018EncoderDecoderAtrous} as the decoder and a ResNet-18 \cite{he2015DeepResidual} as the backbone. Both architectures are adapted from Iakubovskii \cite{Iakubovskii:2019} with PyTorch \cite{paszke2019PyTorchImperative}. All the baseline models inside the teacher are trained with just the segmentation loss described in Equation \ref{eq:l_seg}. 

\textbf{Data Augmentation.} 
To prevent overfitting, we apply the following data augmentation strategy to all training procedures:
\begin{enumerate}
    \itemsep0em     
    \item Random scaling with a scaling factor between 0.5 and 2.0,
    \item Random cropping with the crop size of $768\times768$,
    \item Random horizontal flipping with a flip chance of 50\%. 
\end{enumerate}

\textbf{Training.} 
For all training processes, we employ a \ac{sgd} optimizer based on Robbins and Monro \cite{robbins1951StochasticApproximation} with an initial learning rate of 0.01, momentum of 0.9, and weight decay of 0.0005 as optimizer-specific hyperparameters. In all experiments, the decoder's learning rate is ten times that of the backbone. Additionally, we use polynomial learning rate scheduling to decay the initial learning rate during the training process:
\begin{equation}
lr = lr_{initial} \cdot (1 - \frac{iteration}{total\:iterations})^{0.9},
\end{equation}
where $lr$ is the current learning rate, and $lr_{initial}$ is the initial learning rate. For training the ensemble members, we use the well-known categorical cross-entropy loss described in Equation \ref{eq:l_seg} as the main objective function. In all of the training processes, we train for 200 epochs with a batch size of 16 using mixed precision \cite{micikevicius2018MixedPrecision} on a NVIDIA A100 GPU with 40 GB of memory.

\section{Experiments}\label{sec: results}

\begin{table*}[ht!]
\begin{center}
\begin{adjustbox}{width=\textwidth}
\setlength\extrarowheight{1mm}
\begin{tabular}{l|*{19}{c}|c}
& \rotatebox{90}{Road} & \rotatebox{90}{Sidewalk} & \rotatebox{90}{Building} & \rotatebox{90}{Wall} & \rotatebox{90}{Fence} & \rotatebox{90}{Pole} & \rotatebox{90}{Tr. Light} & \rotatebox{90}{Tr. Sign} & \rotatebox{90}{Vegetation} & \rotatebox{90}{Terrain} & \rotatebox{90}{Sky} & \rotatebox{90}{Person} & \rotatebox{90}{Rider} & \rotatebox{90}{Car} & \rotatebox{90}{Truck} & \rotatebox{90}{Bus} & \rotatebox{90}{Train} & \rotatebox{90}{Motorbike} & \rotatebox{90}{Bicycle} & \cellcolor{gray!20}\rotatebox{90}{Mean} \\ \hline
Teacher IoU \cite{holder2021EfficientUncertainty} & 96.1 & 79.4 & 91.4 & 43.2 & 56.3 & 58.4 & 62.0 & 73.0 & 91.7 & 59.6 & 93.7 & 78.2 & 55.3 & 93.5 & 66.8 & 79.3 & 67.7 & 53.4 & 74.3 &  \cellcolor{gray!20}72.3  \\
Student IoU \cite{holder2021EfficientUncertainty} & 96.4 & 77.2 & 90.0 & 42.6 & 54.7 & 47.6 & 51.1 & 65.2 & 90.4 & 56.4 & 91.9 & 73.9 & 49.8 & 92.1 & 61.7 & 72.3 & 62.5 & 49.3 & 68.9 &  \cellcolor{gray!20}68.1 \\ \hline
Teacher IoU (Ours) & 97.8 & 82.2 & 90.7 & 50.4 & 54.5 & 54.9 & 57.8 & 69.3 & 91.5 & 62.7 & 94.3 & 75.4 & 53.5 & 93.2 & 69.6 & 75.9 & 64.0 & 47.6 & 69.5 &  \cellcolor{gray!20}71.3  \\
Student IoU (Ours) & 98.0 & 83.5 & 91.4 & 46.7 & 55.7 & 59.1 & 63.3 & 73.3 & 91.8 & 63.1 & 94.2 & 79.0 & 57.9 & 93.9 & 74.7 & 83.8 & 69.4 & 50.1 & 73.6 &  \cellcolor{gray!20}73.8 \\ \hline \hline
Difference \cite{holder2021EfficientUncertainty} $\uparrow$ & \textbf{0.3} & {-2.2} & {-1.4} & \textbf{-0.6} & {-1.6} & {-10.8} & {-10.9} & {-7.8} & {-1.3} & {-3.2} & { -1.8} & {-4.3} & {-5.5} & {-1.4} & {-5.1} & {-7.0} & {-5.2} & {-4.1} & {-5.4} &  \cellcolor{gray!20}{-4.2}  \\ \hline
Difference (Ours) $\uparrow$ & {0.2} & \textbf{1.3} & \textbf{0.7} & {-3.7} & \textbf{1.2} & \textbf{4.2} & \textbf{5.5} & \textbf{4.0} & \textbf{0.3} & \textbf{0.4} & \textbf{ -0.1} & \textbf{3.6} & \textbf{4.4} & \textbf{0.7} & \textbf{5.1} & \textbf{7.9} & \textbf{5.4} & \textbf{2.5} & \textbf{4.1} &  \cellcolor{gray!20}\textbf{2.5}  \\ \hline
\end{tabular}
\end{adjustbox}
\end{center}
\caption{Quantitative comparison between the student's and the teacher's class-wise Intersection over Union (IoU). Higher IoU values denote better segmentation results, which are preferred. For the difference, the teacher's results are subtracted from the student's results.}
\label{table: class_wise_iou}
\end{table*}

\begin{table*}[ht!]
\begin{center}
\begin{adjustbox}{width=\textwidth}
\setlength\extrarowheight{1mm}
\begin{tabular}{l|*{19}{c}|c}
& \rotatebox{90}{Road} & \rotatebox{90}{Sidewalk} & \rotatebox{90}{Building} & \rotatebox{90}{Wall} & \rotatebox{90}{Fence} & \rotatebox{90}{Pole} & \rotatebox{90}{Tr. Light} & \rotatebox{90}{Tr. Sign} & \rotatebox{90}{Vegetation} & \rotatebox{90}{Terrain} & \rotatebox{90}{Sky} & \rotatebox{90}{Person} & \rotatebox{90}{Rider} & \rotatebox{90}{Car} & \rotatebox{90}{Truck} & \rotatebox{90}{Bus} & \rotatebox{90}{Train} & \rotatebox{90}{Motorbike} & \rotatebox{90}{Bicycle} & \cellcolor{gray!20}\rotatebox{90}{Mean} \\ \hline
Teacher Unc. \cite{holder2021EfficientUncertainty} & 0.029 & 0.097 & 0.055 & 0.210 & 0.147 & 0.100 & 0.128 & 0.108 & 0.028 & 0.129 & 0.030 & 0.068 & 0.125 & 0.030 & 0.176 & 0.155 & 0.257 & 0.165 & 0.082 &  \cellcolor{gray!20}0.111\\
Student Unc. \cite{holder2021EfficientUncertainty} & 0.032 & 0.086 & 0.077 & 0.155 & 0.141 & 0.133 & 0.135 & 0.111 & 0.055 & 0.127 & 0.046 & 0.097 & 0.127 & 0.046 & 0.108 & 0.100 & 0.127 & 0.135 & 0.130 & \cellcolor{gray!20}0.104\\\hline
Teacher Unc. (Ours) & 0.024 & 0.064 & 0.038 & 0.150 & 0.165 & 0.100 & 0.142 & 0.102 & 0.025 & 0.101 & 0.031 & 0.109 & 0.120 & 0.043 & 0.175 & 0.163 & 0.195 & 0.158 & 0.108 & \cellcolor{gray!20}0.106 \\
Student Unc. (Ours) & 0.018 & 0.065 & 0.035 & 0.144 & 0.160 & 0.128 & 0.112 & 0.097 & 0.027 & 0.126 & 0.025 & 0.117 & 0.105 & 0.038 & 0.200 & 0.144 & 0.171 & 0.190 & 0.150 & \cellcolor{gray!20}0.108\\\hline \hline
Difference \cite{holder2021EfficientUncertainty} $\downarrow$ & \cellcolor{green!60}{0.003} & \cellcolor{green!40}{-0.011} & \cellcolor{green!20}{0.022} & \cellcolor{red!40}{-0.055} & \cellcolor{green!30}{-0.006} & {0.033} & \cellcolor{green!60}{0.007} & \cellcolor{green!60}{0.003} & \cellcolor{green!20}{0.027} & \cellcolor{green!60}{0.002} & \cellcolor{green!40}{0.016} & \cellcolor{green!20}{0.029} & \cellcolor{green!60}{0.002} & \cellcolor{green!40}{0.016} & \cellcolor{red!60}{-0.068} & \cellcolor{red!40}{-0.055} & \cellcolor{red!60}{-0.130} & \cellcolor{green!20}{-0.030} & \cellcolor{red!20}{0.048} &  \cellcolor{gray!20}{-0.007}  \\ \hline
Difference (Ours) $\downarrow$ & \cellcolor{green!60}{-0.006} & \cellcolor{green!60}{0.001} & \cellcolor{green!60}{-0.003} & \cellcolor{green!60}{-0.006} & \cellcolor{green!60}{-0.005} & \cellcolor{green!20}{0.028} & \cellcolor{green!20}{-0.03} & \cellcolor{green!60}{-0.005} & \cellcolor{green!60}{0.002} & \cellcolor{green!20}{0.025} & \cellcolor{green!60}{-0.006} & \cellcolor{green!60}{0.008} & \cellcolor{green!40}{-0.015} & \cellcolor{green!60}{-0.005} & \cellcolor{green!20}{0.025} & \cellcolor{green!40}{-0.019} & \cellcolor{green!20}{-0.024} & {0.032} & \cellcolor{red!20}{0.042} &  \cellcolor{gray!20}{0.002}  \\ \hline
\end{tabular}
\end{adjustbox}
\end{center}
\caption{Quantitative comparison between the student's and the teacher's class-wise predictive uncertainties. In this case, a smaller difference is preferred as the student is trained to predict the same uncertainties as the teacher. The differences are calculated by subtracting the teacher's results from the student's results. They are highlighted based on the absolute differences being: \colorbox{green!60}{$\leq 0.01$}, \colorbox{green!40}{$\leq 0.02$}, \colorbox{green!20}{$\leq 0.03$}, {$\leq 0.04$}, \colorbox{red!20}{$\leq 0.05$}, \colorbox{red!40}{$\leq 0.06$}, \colorbox{red!60}{$\geq 0.06$}.}
\label{table: class_wise_uncertainty}
\end{table*}

In this section, we demonstrate a variety of experiments conducted on the basis of the Cityscapes dataset to manifest the value of \ac{dudes}. Firstly, we compare the inference time and class-wise uncertainties between the teacher and the student model. Secondly, we evaluate the student's predictions qualitatively. Thirdly, we provide ablation studies.

\subsection{Dataset}
All of our experiments are based on the Cityscapes dataset \cite{cordts2016CityscapesDataset}, a freely available urban street scene dataset. It consists of \SI{2975} training images, \SI{500} validation images, and \SI{1525} test images. Since the test images are not publicly available, we use the validation images for testing in all of our experiments. Each RGB image is $2048\times1024$ in size, with each pixel assigned to one of 19 class labels or a void label. The void ground truth pixels are excluded during training and evaluation in the segmentation task, but they are used to evaluate the uncertainty outputs as they indicate the model's ability to distinguish between in-domain and out-of-domain samples.

\subsection{Quantitative Evaluation}
\begin{figure*}[ht!]
    \centering
    \begin{subfigure}{0.195\textwidth}
        \includegraphics[width=\textwidth]{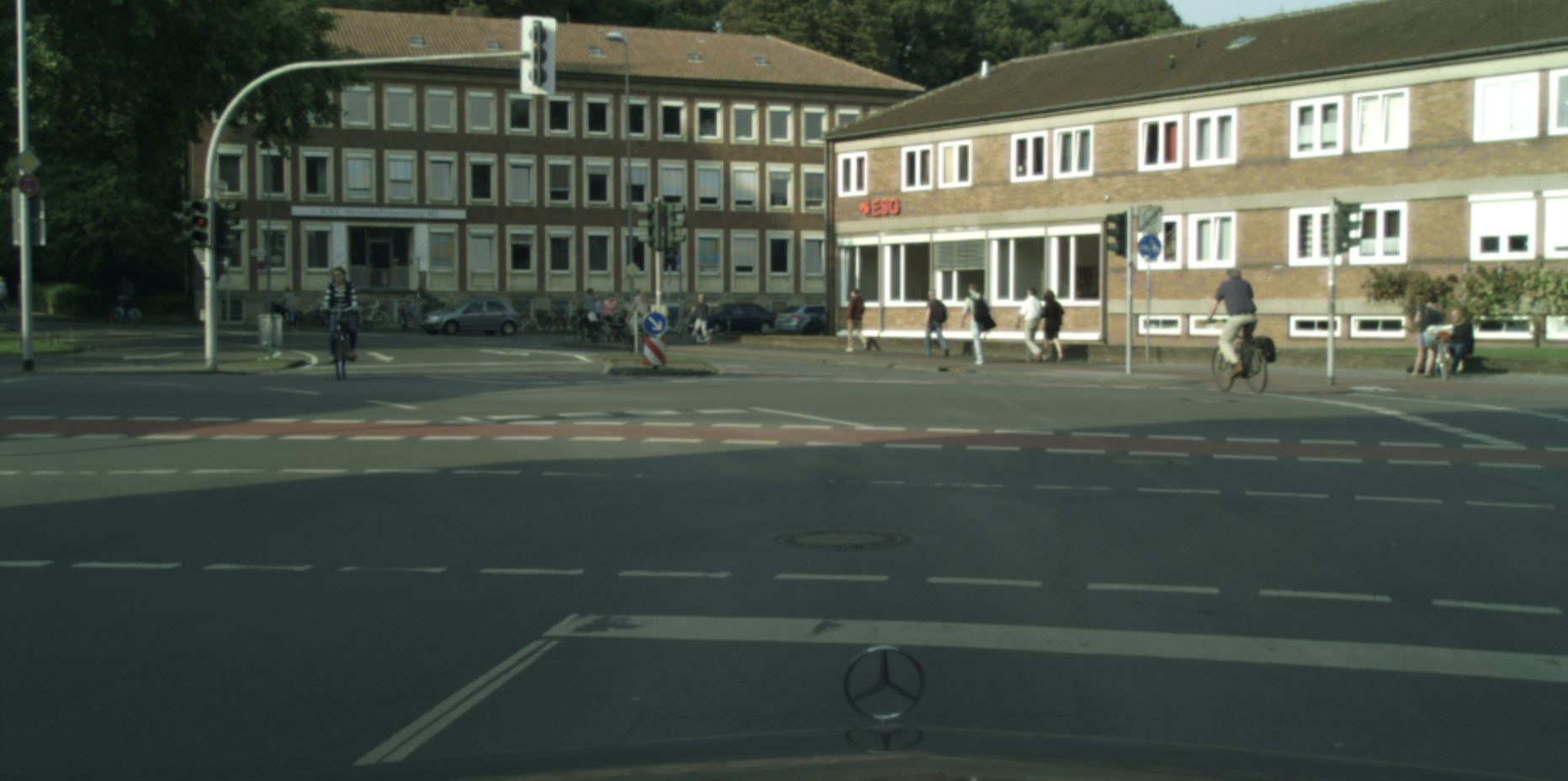}
    \end{subfigure}
    \begin{subfigure}{0.195\textwidth}
        \includegraphics[width=\textwidth]{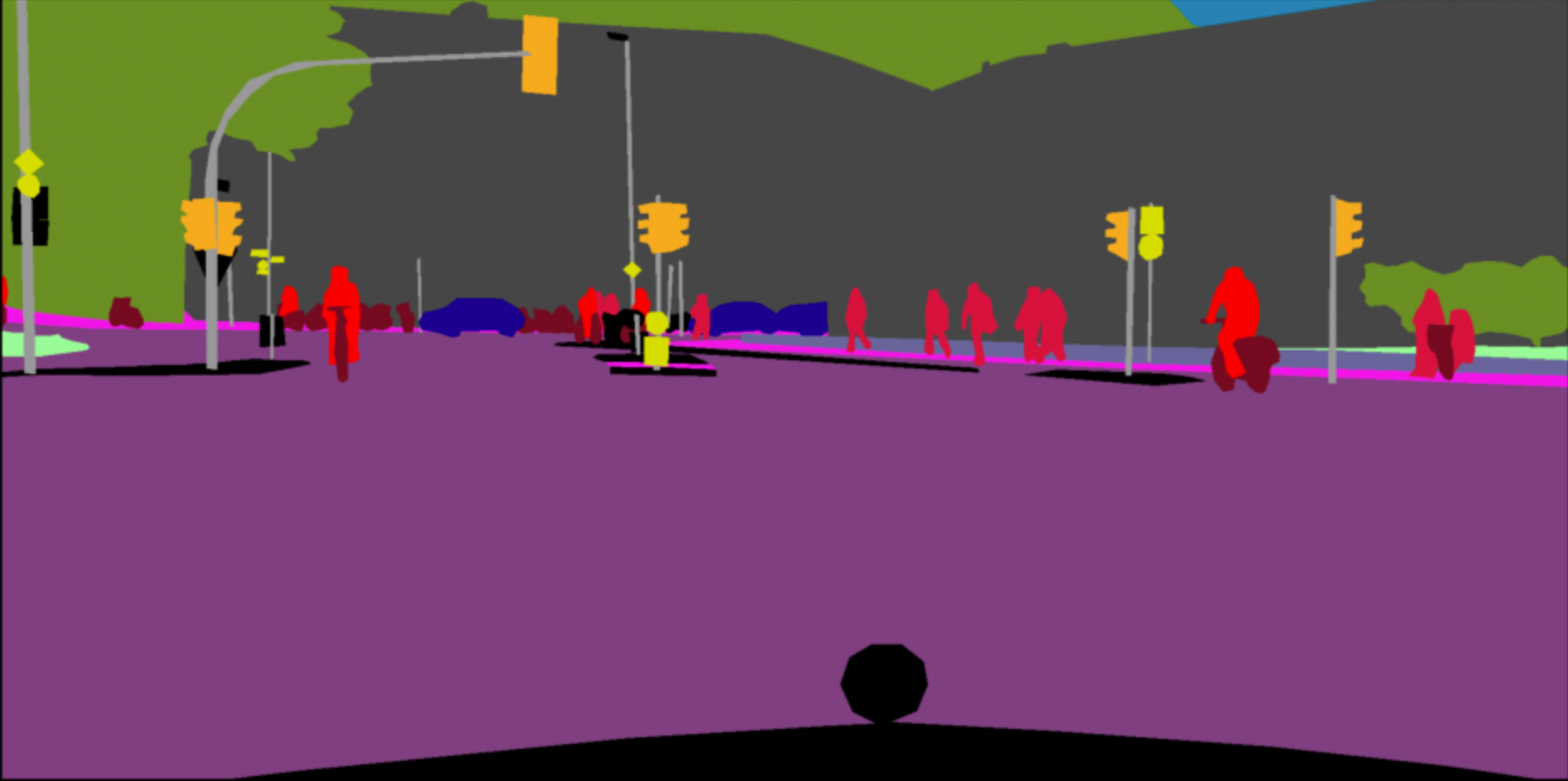}
    \end{subfigure}
    \begin{subfigure}{0.195\textwidth}
        \includegraphics[width=\textwidth]{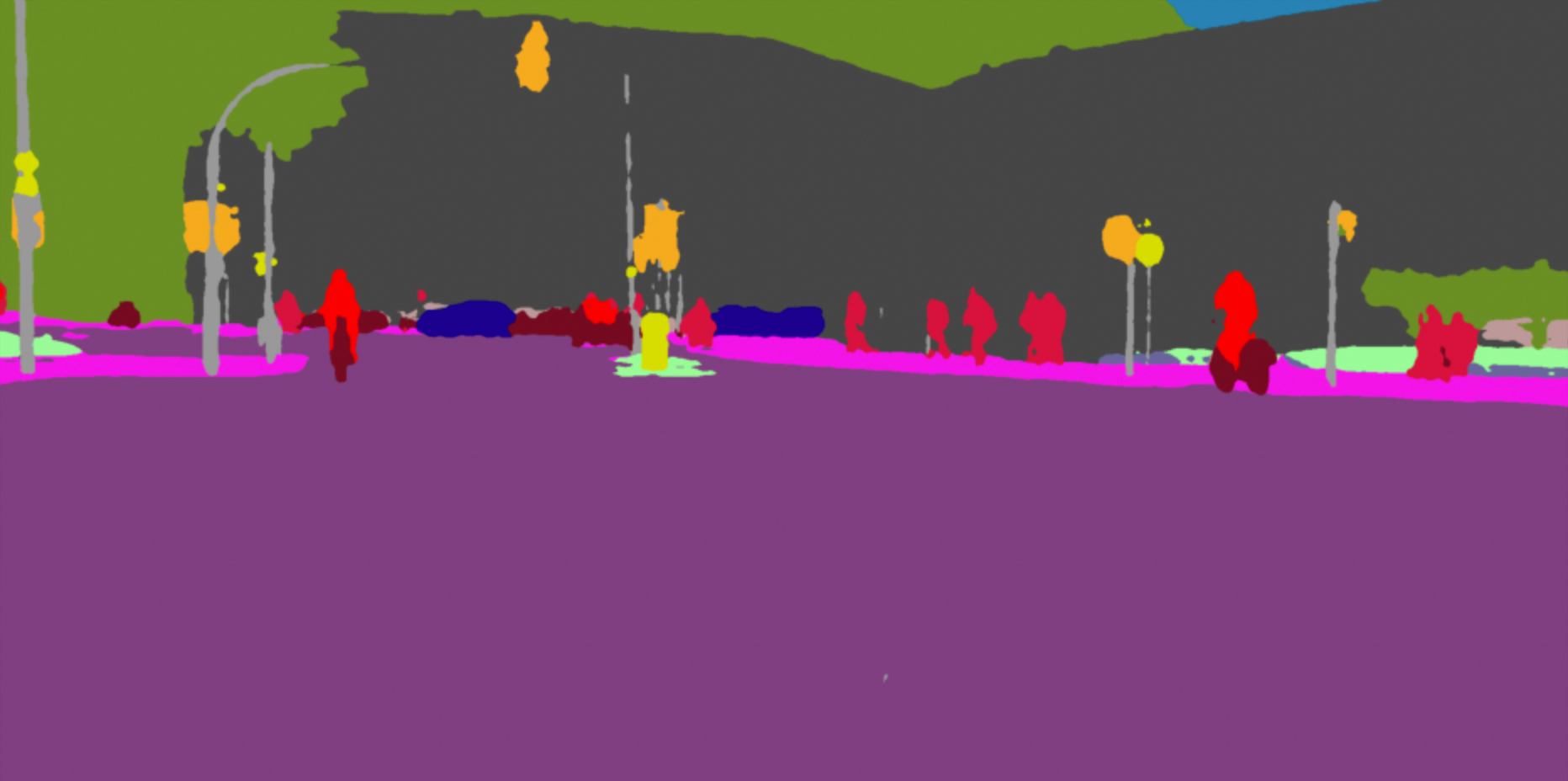}
    \end{subfigure}
    \begin{subfigure}{0.195\textwidth}
        \includegraphics[width=\textwidth]{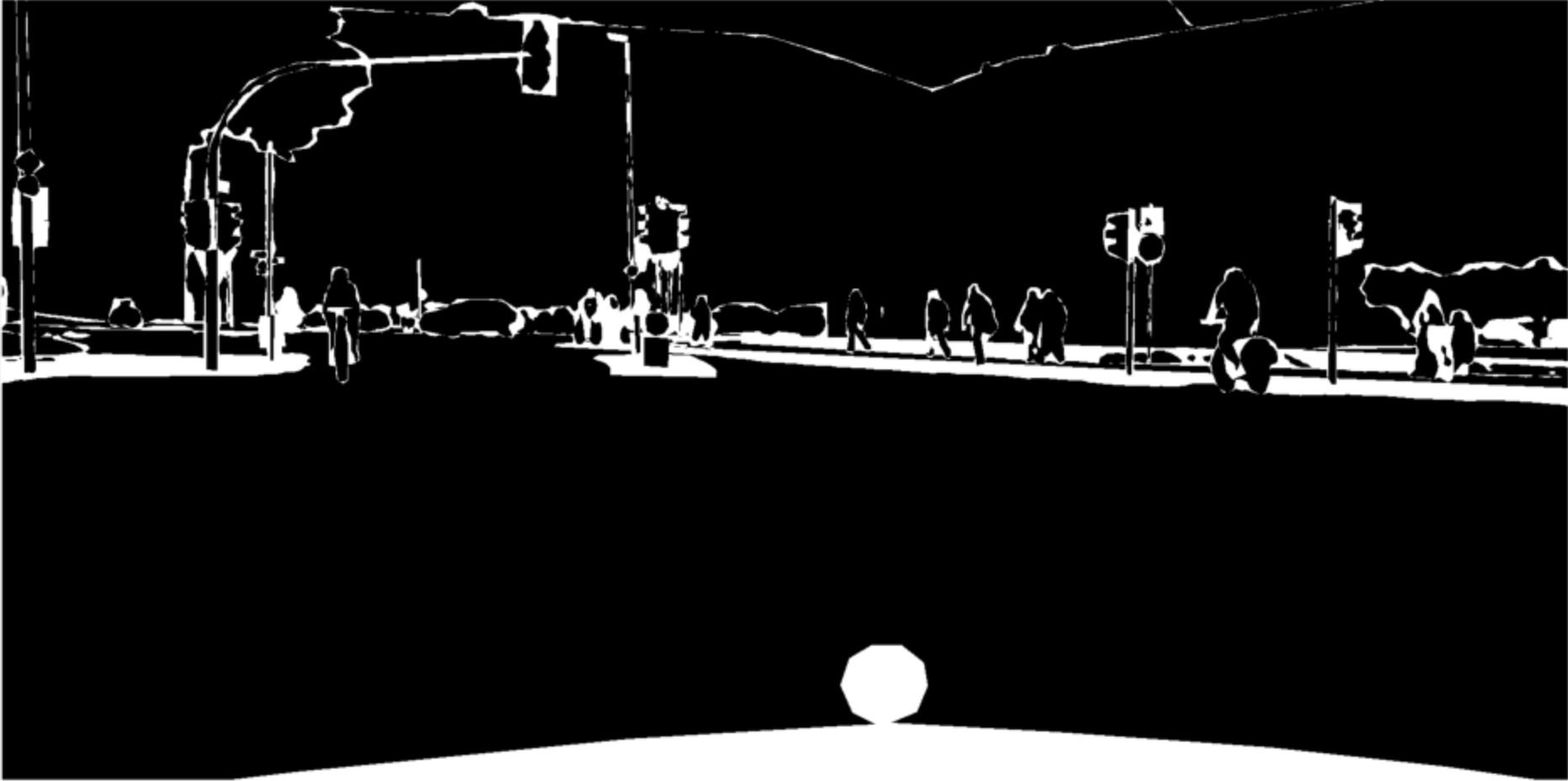}
    \end{subfigure}
    \begin{subfigure}{0.195\textwidth}
        \includegraphics[width=\textwidth]{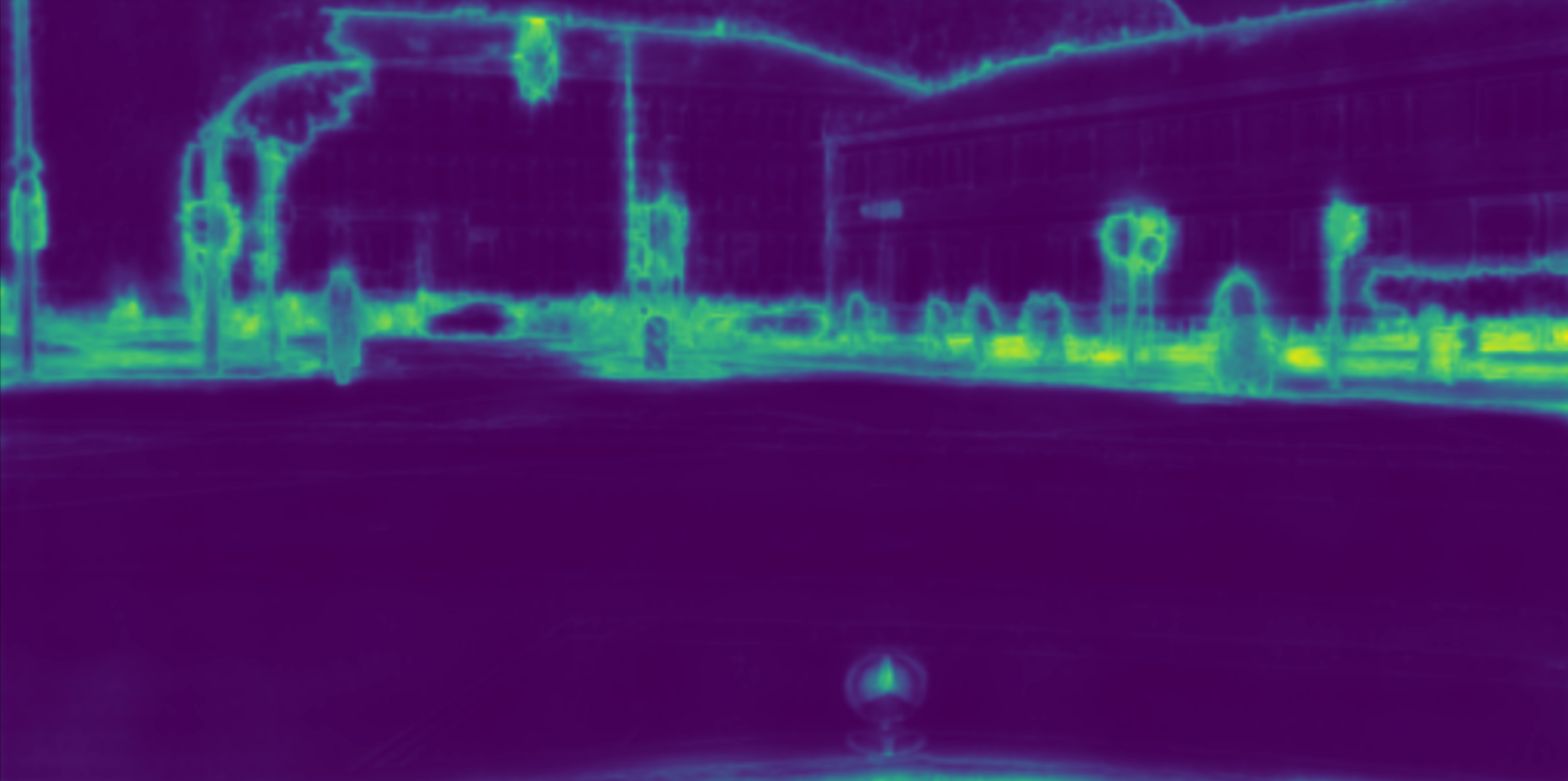}
    \end{subfigure}
    \begin{subfigure}{0.195\textwidth}
        \includegraphics[width=\textwidth]{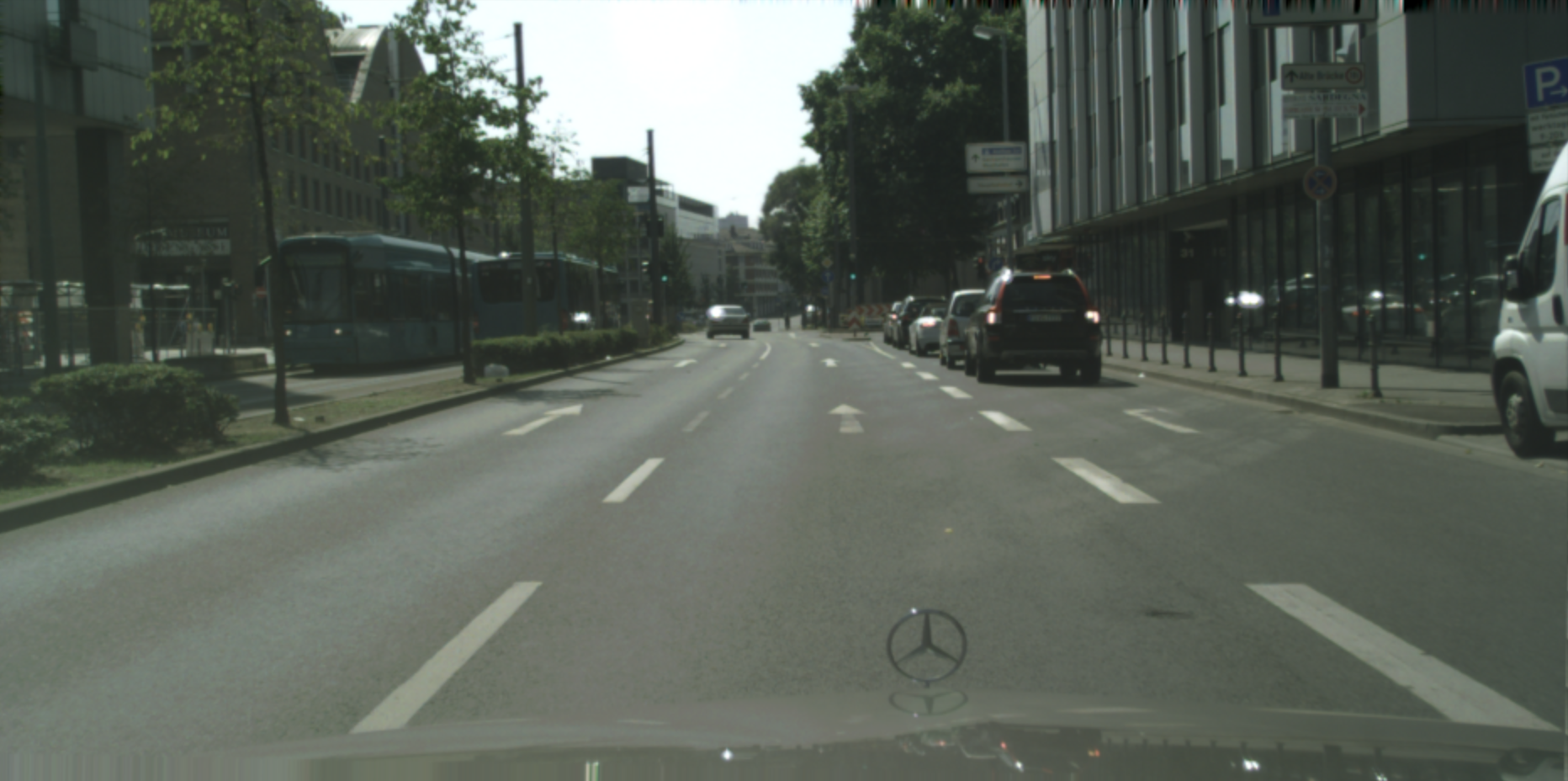}
    \end{subfigure}
    \begin{subfigure}{0.195\textwidth}
        \includegraphics[width=\textwidth]{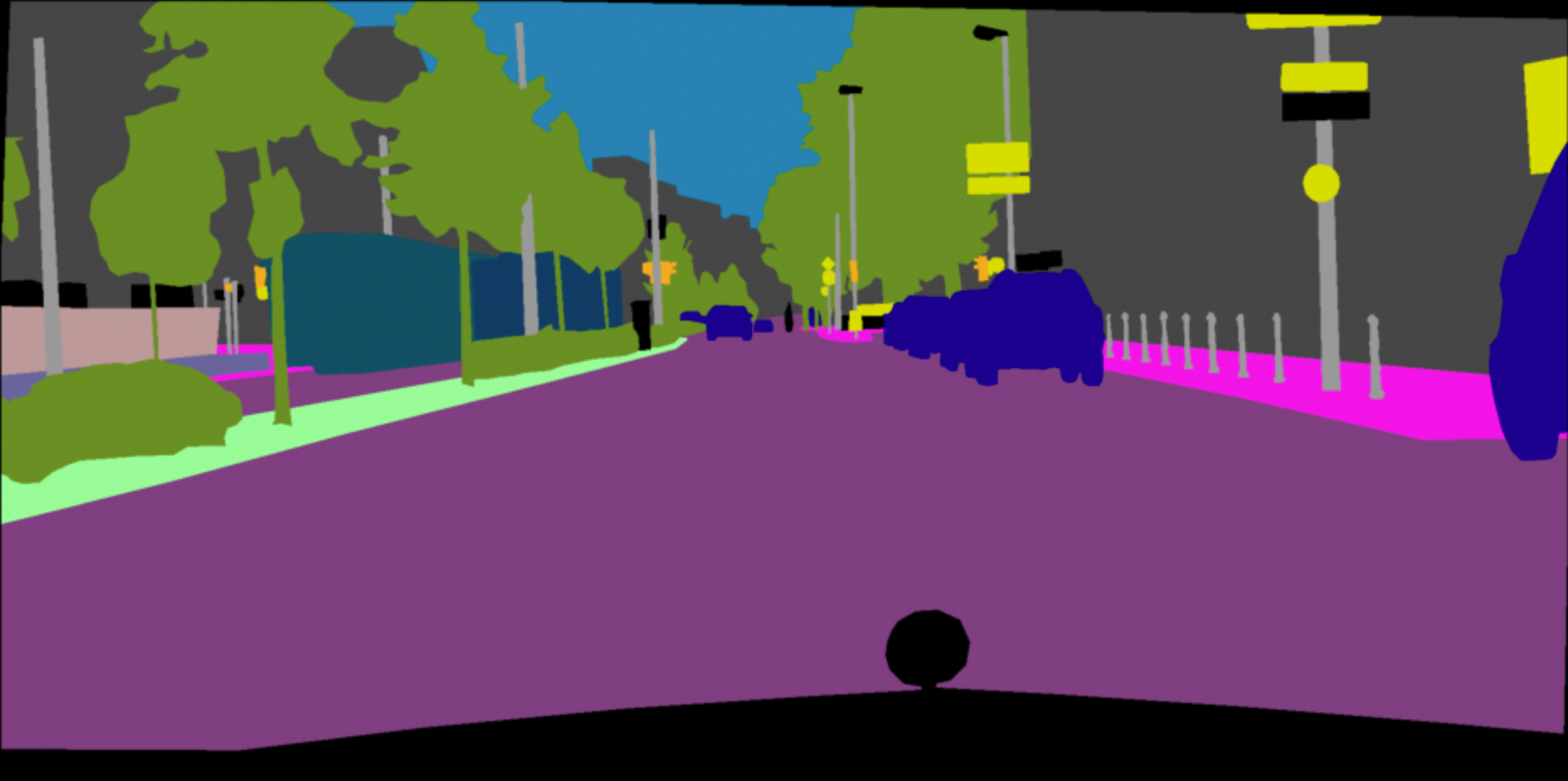}
    \end{subfigure}
    \begin{subfigure}{0.195\textwidth}
        \includegraphics[width=\textwidth]{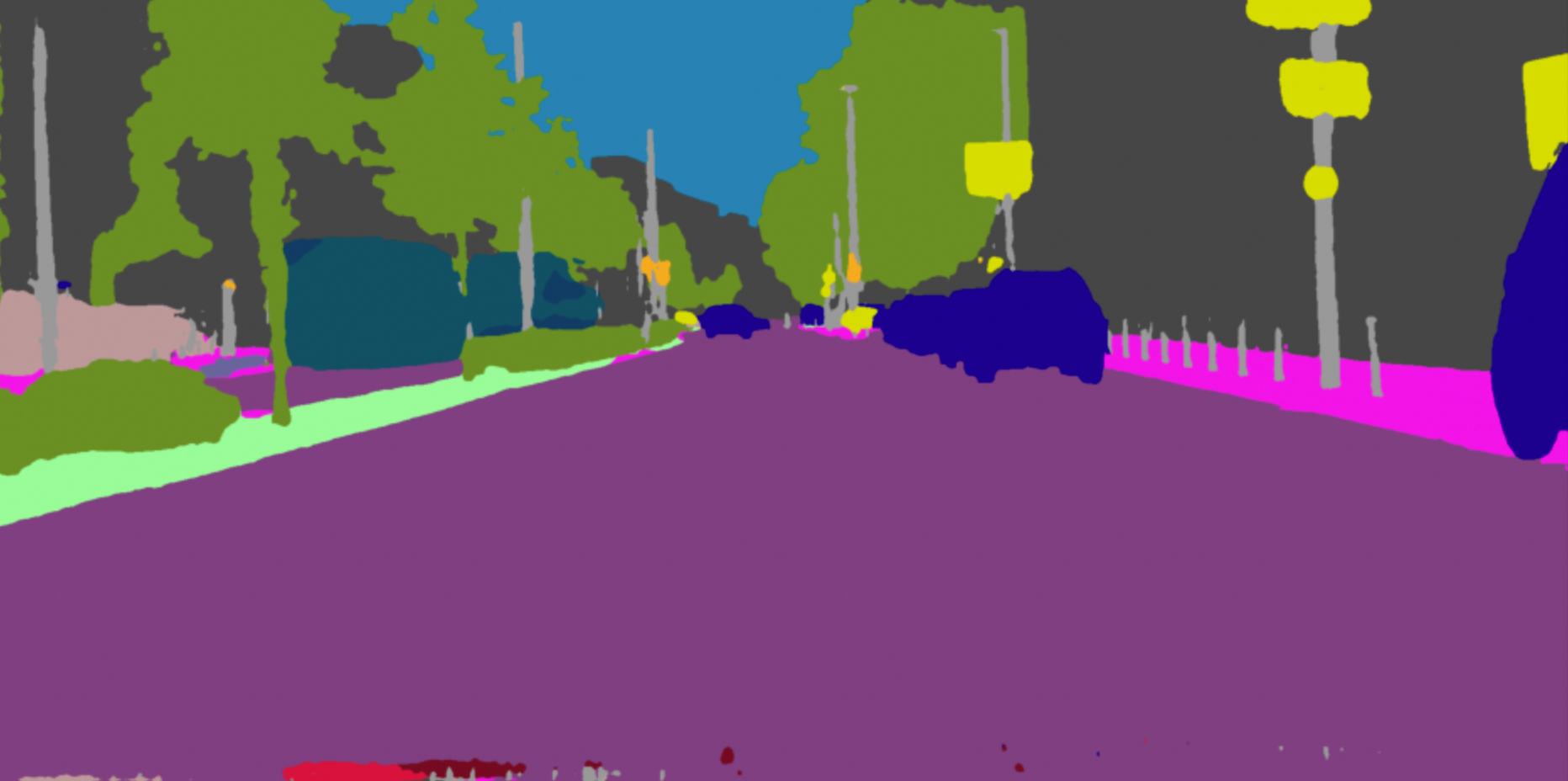}
    \end{subfigure}
    \begin{subfigure}{0.195\textwidth}
        \includegraphics[width=\textwidth]{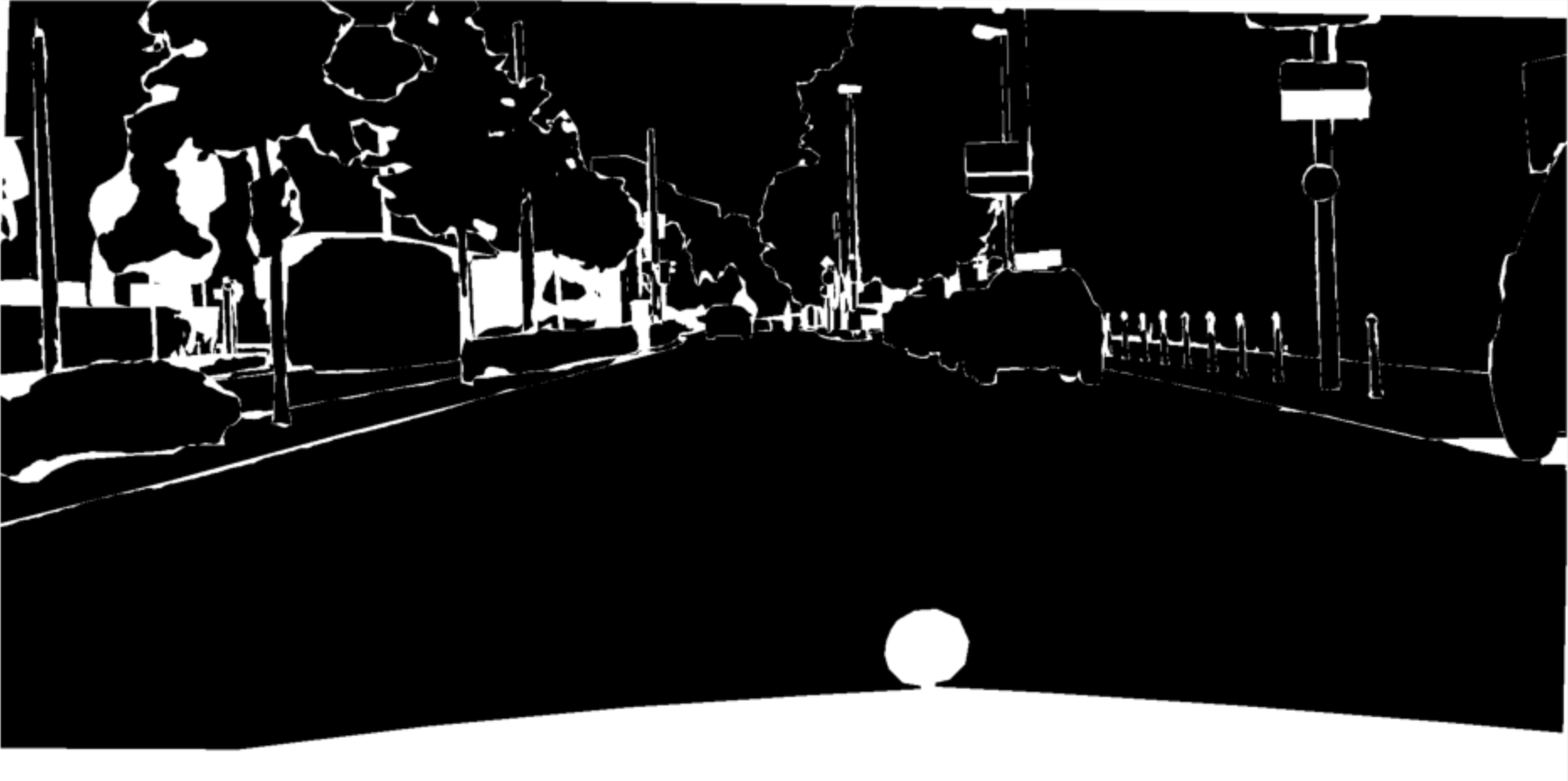}
    \end{subfigure}
    \begin{subfigure}{0.195\textwidth}
        \includegraphics[width=\textwidth]{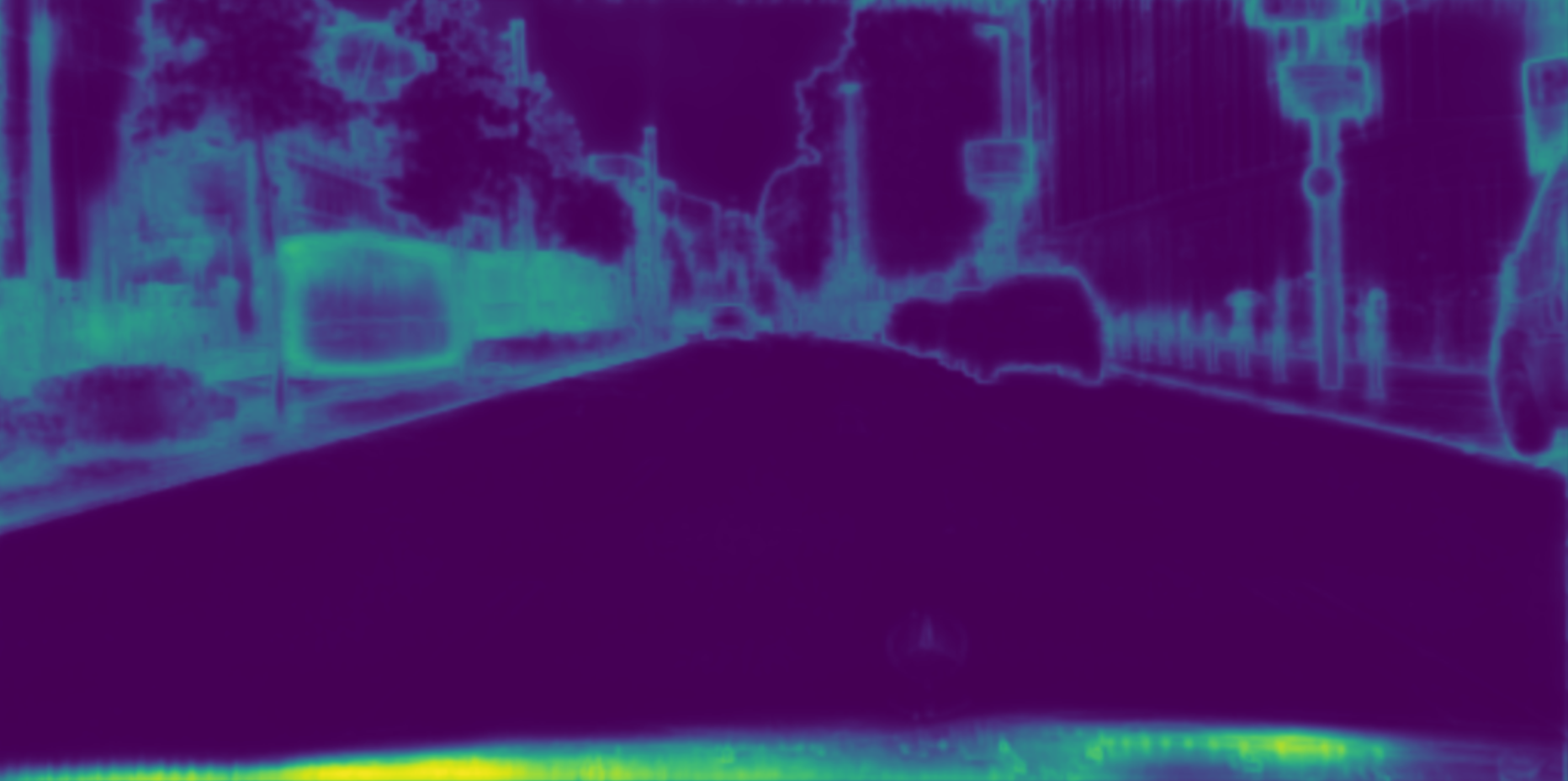}
    \end{subfigure}
    \begin{subfigure}{0.195\textwidth}
        \includegraphics[width=\textwidth]{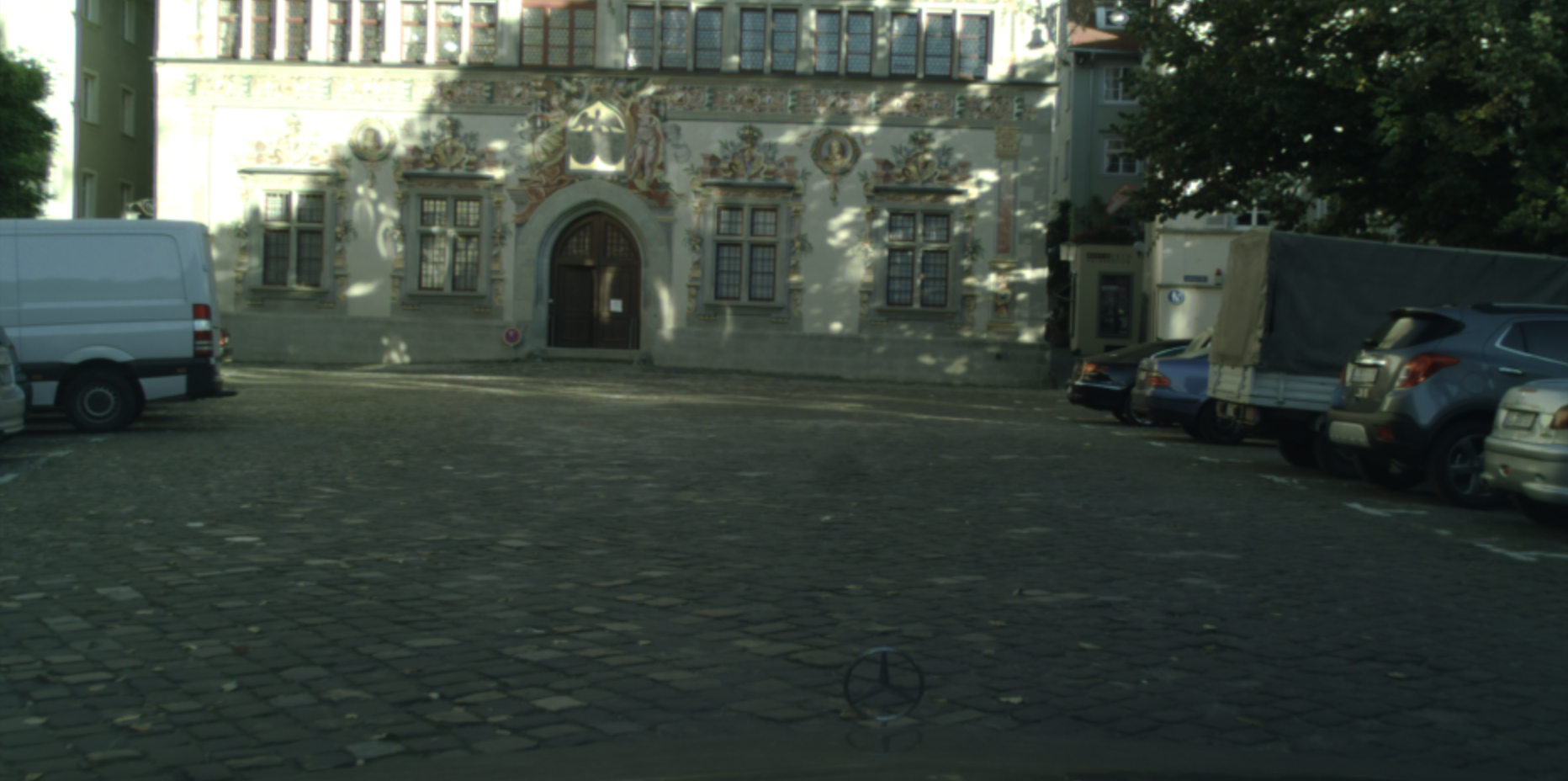}
    \end{subfigure}
    \begin{subfigure}{0.195\textwidth}
        \includegraphics[width=\textwidth]{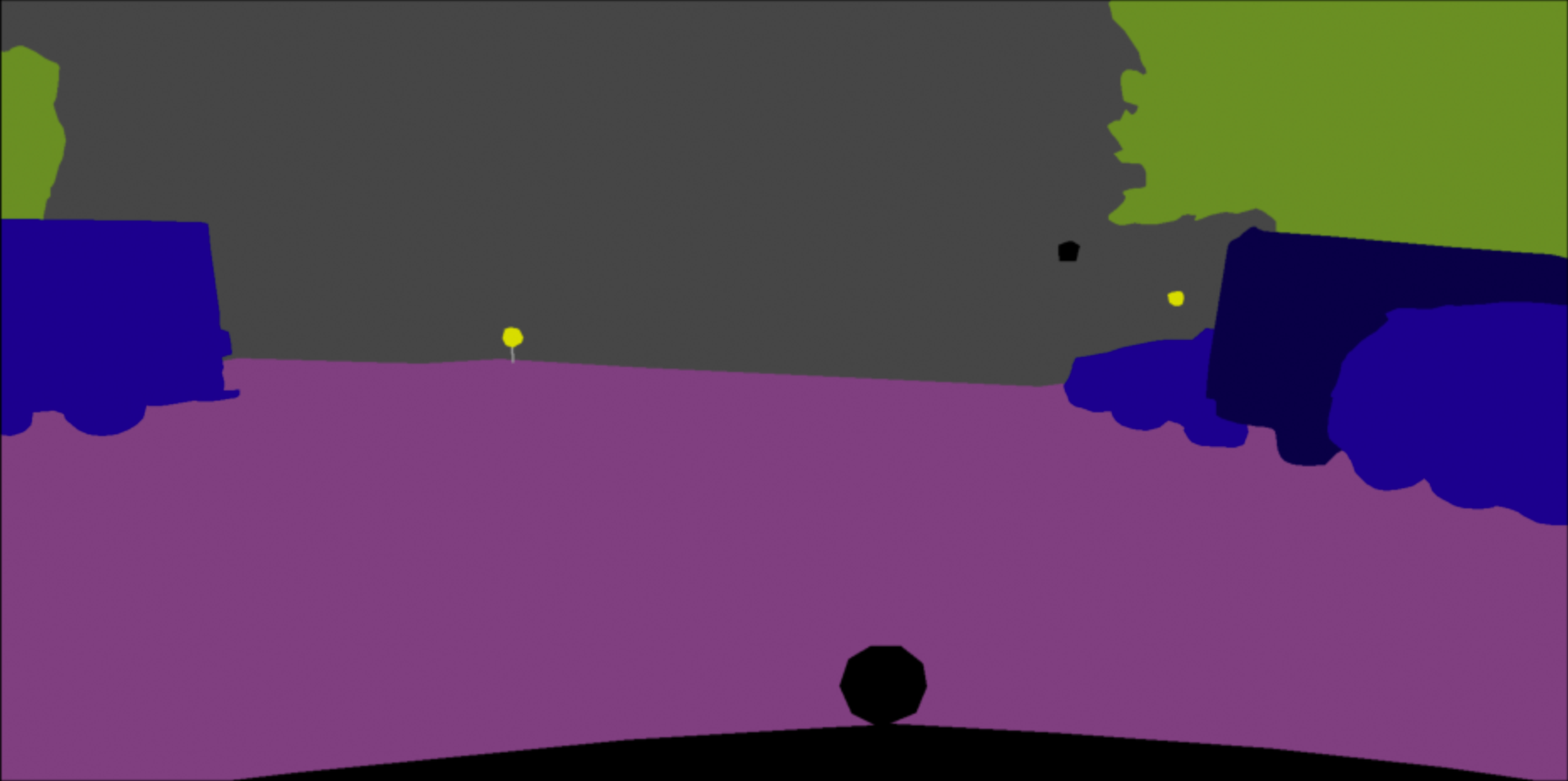}
    \end{subfigure}
    \begin{subfigure}{0.195\textwidth}
        \includegraphics[width=\textwidth]{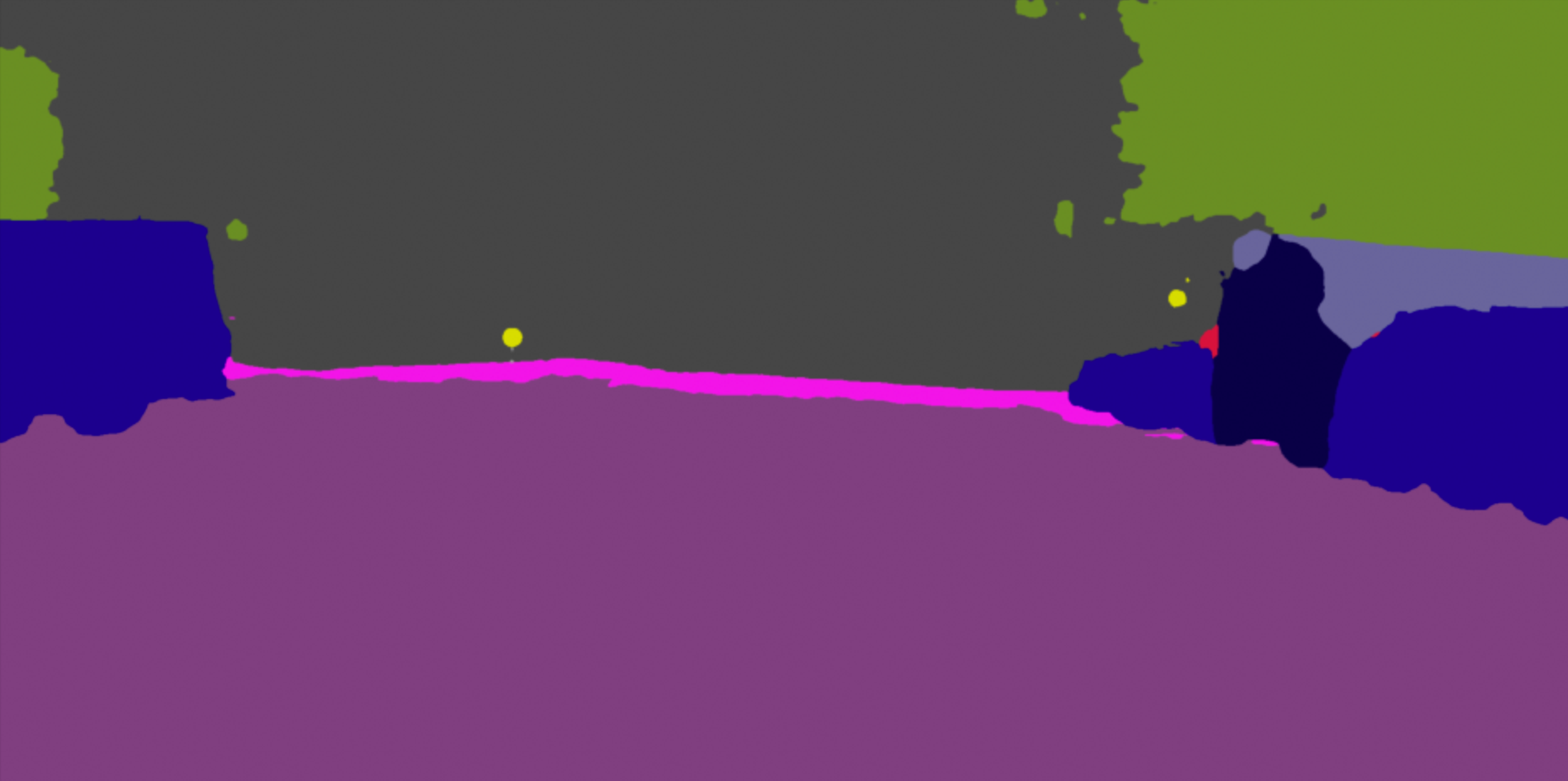}
    \end{subfigure}
    \begin{subfigure}{0.195\textwidth}
        \includegraphics[width=\textwidth]{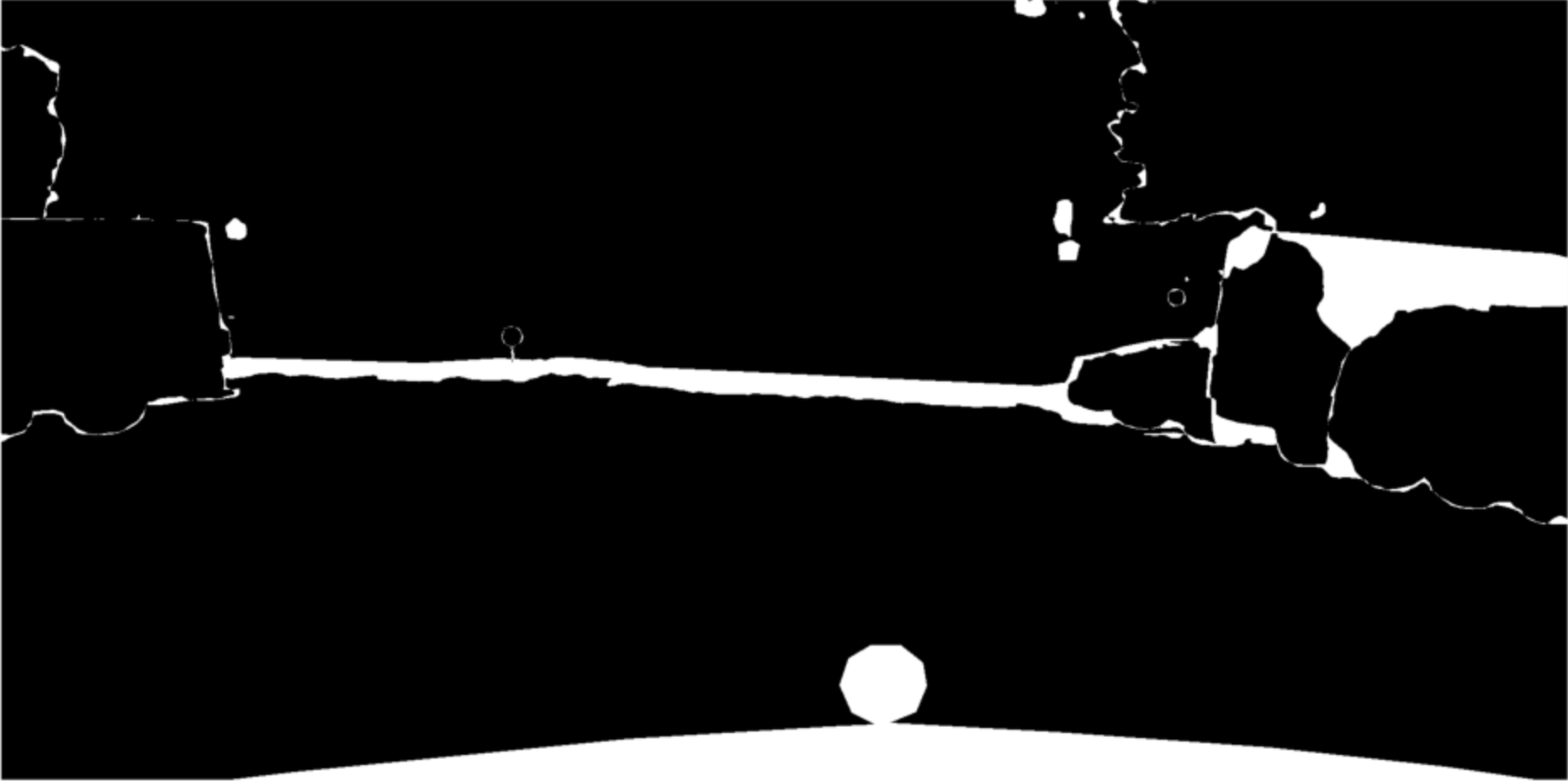}
    \end{subfigure}
    \begin{subfigure}{0.195\textwidth}
        \includegraphics[width=\textwidth]{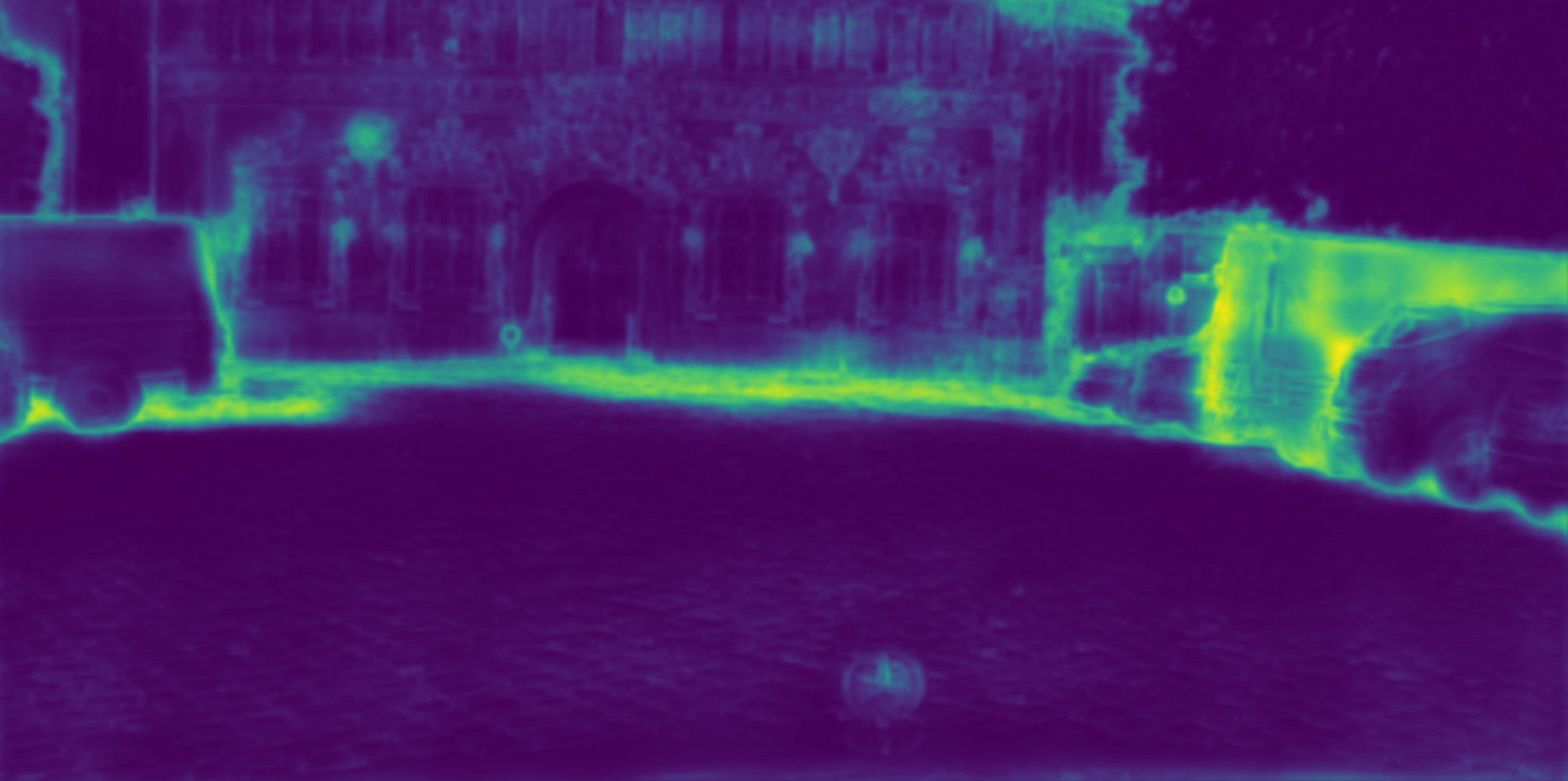}
    \end{subfigure}
    \begin{subfigure}{0.195\textwidth}
        \includegraphics[width=\textwidth]{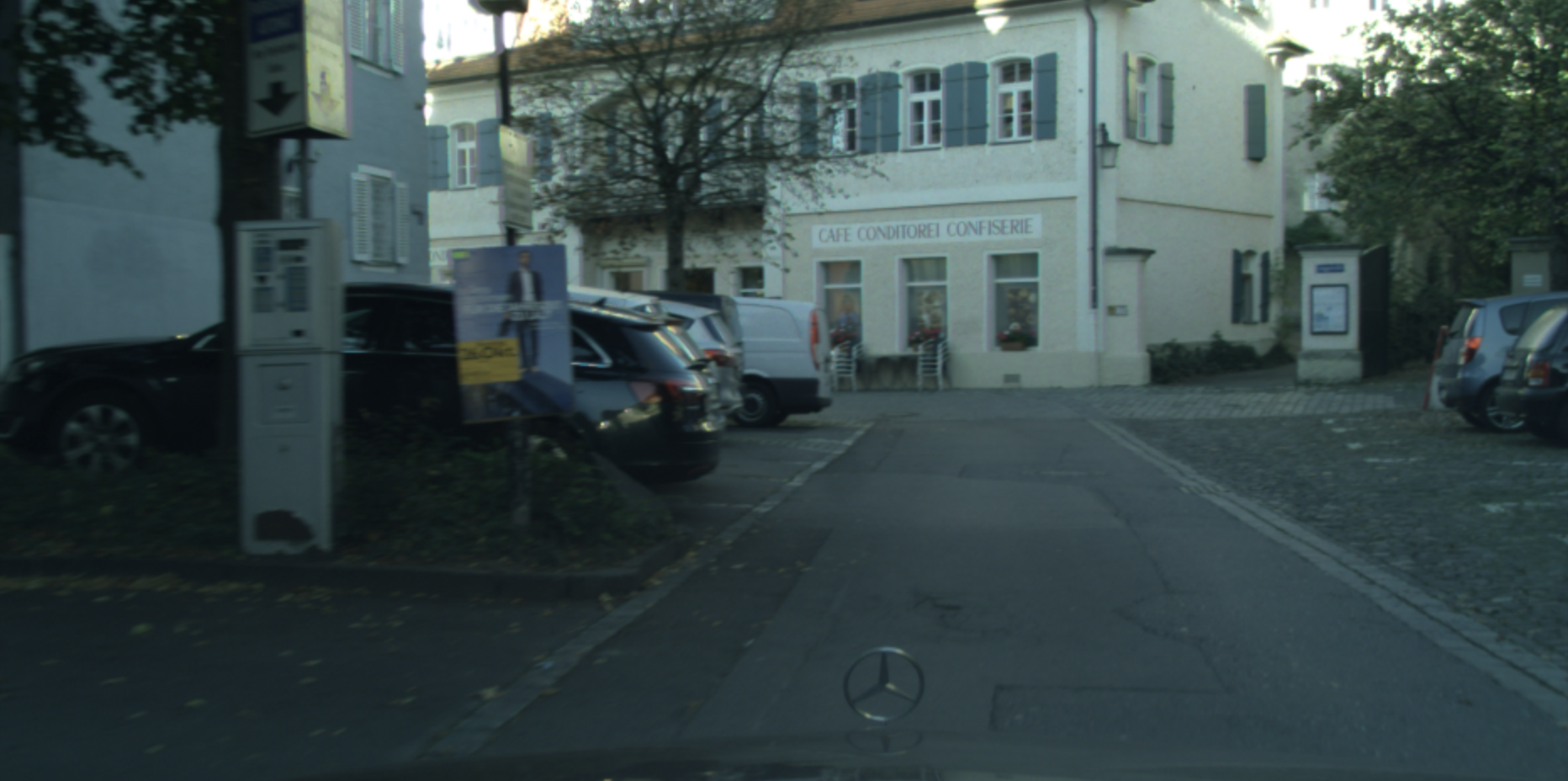}
        \caption{Image}
    \end{subfigure}
    \begin{subfigure}{0.195\textwidth}
        \includegraphics[width=\textwidth]{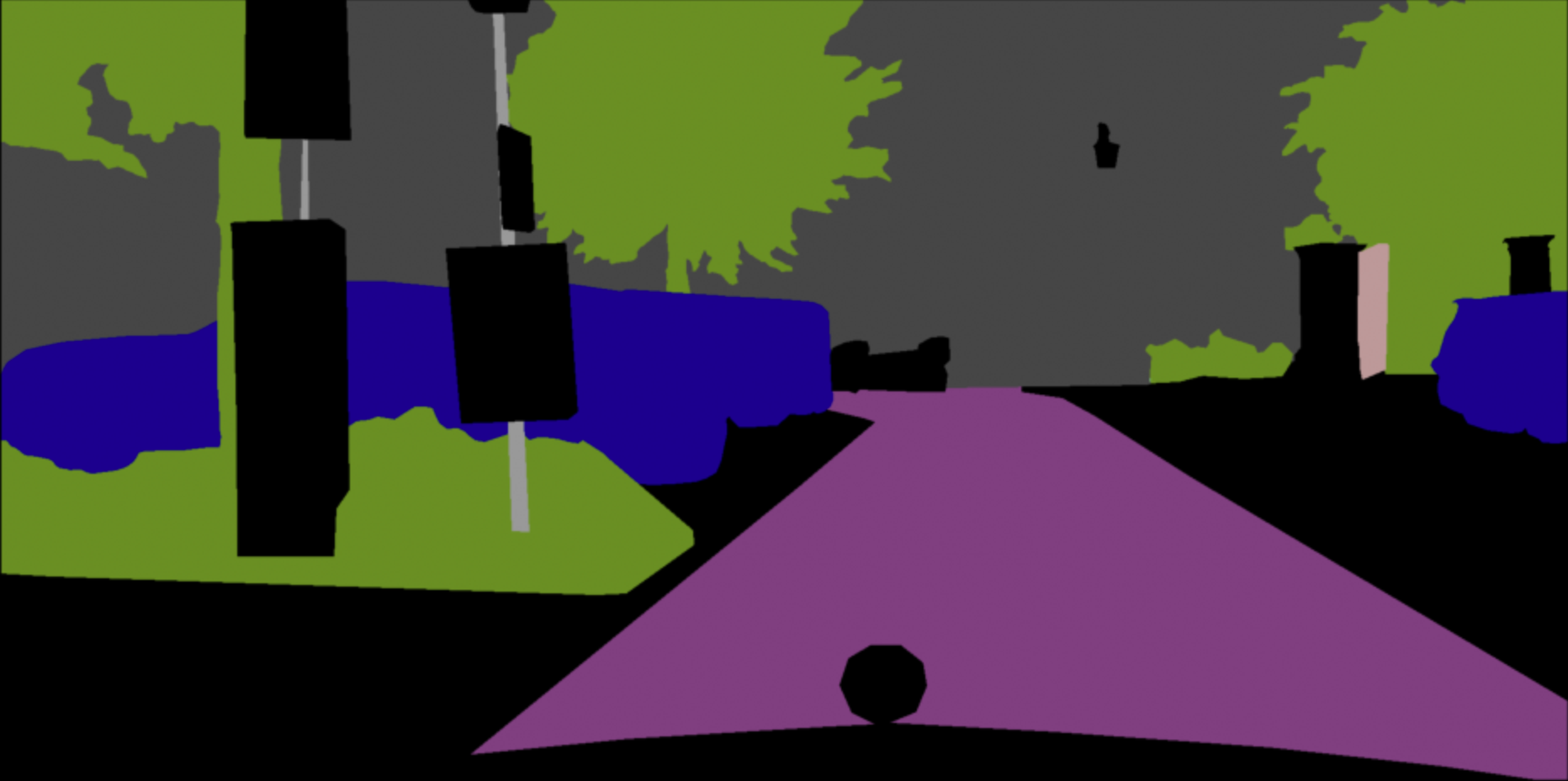}
        \caption{Ground Truth Label}
    \end{subfigure}
    \begin{subfigure}{0.195\textwidth}
        \includegraphics[width=\textwidth]{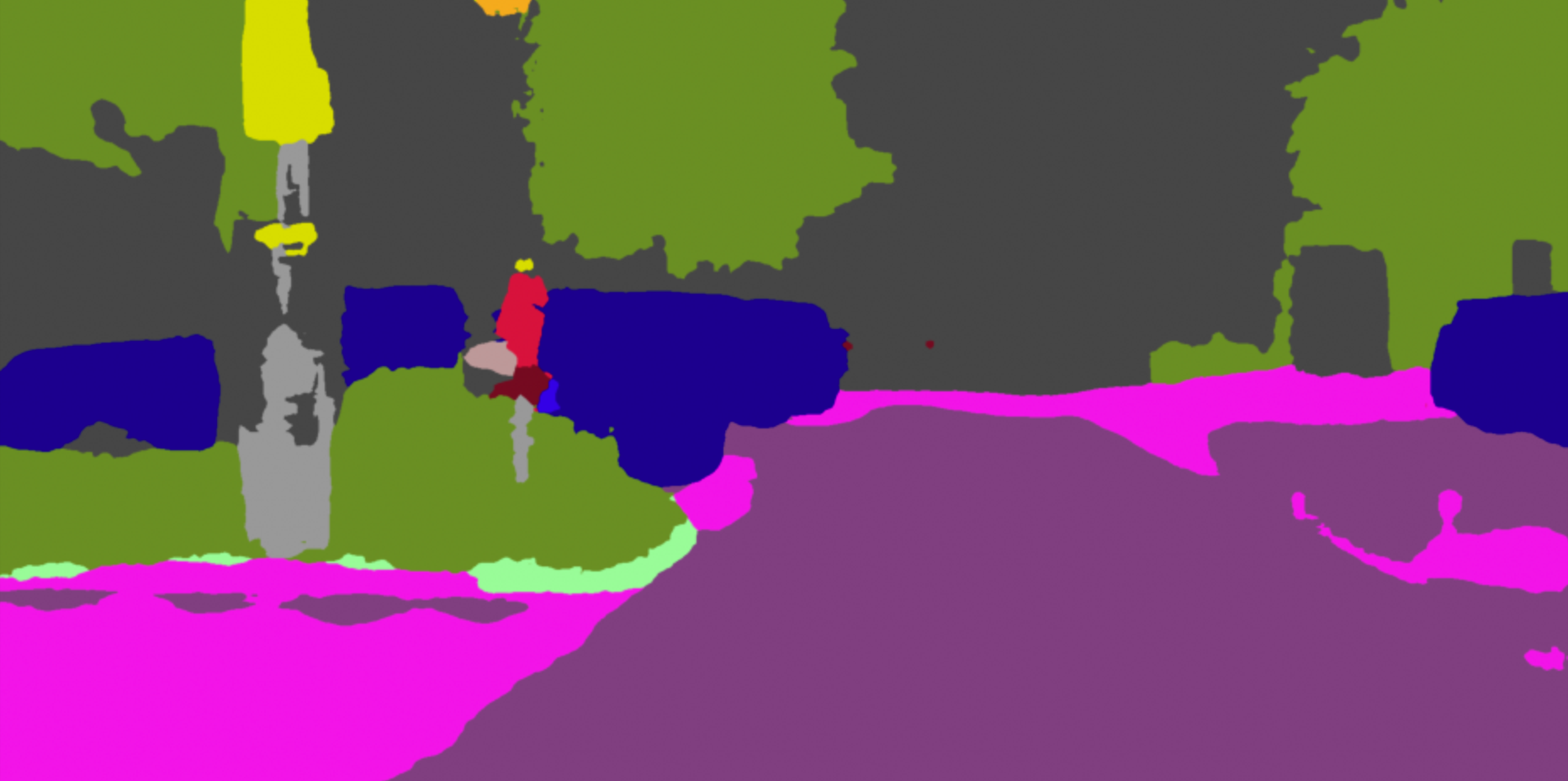}
        \caption{Segmentation Prediction}
    \end{subfigure}
    \begin{subfigure}{0.195\textwidth}
        \includegraphics[width=\textwidth]{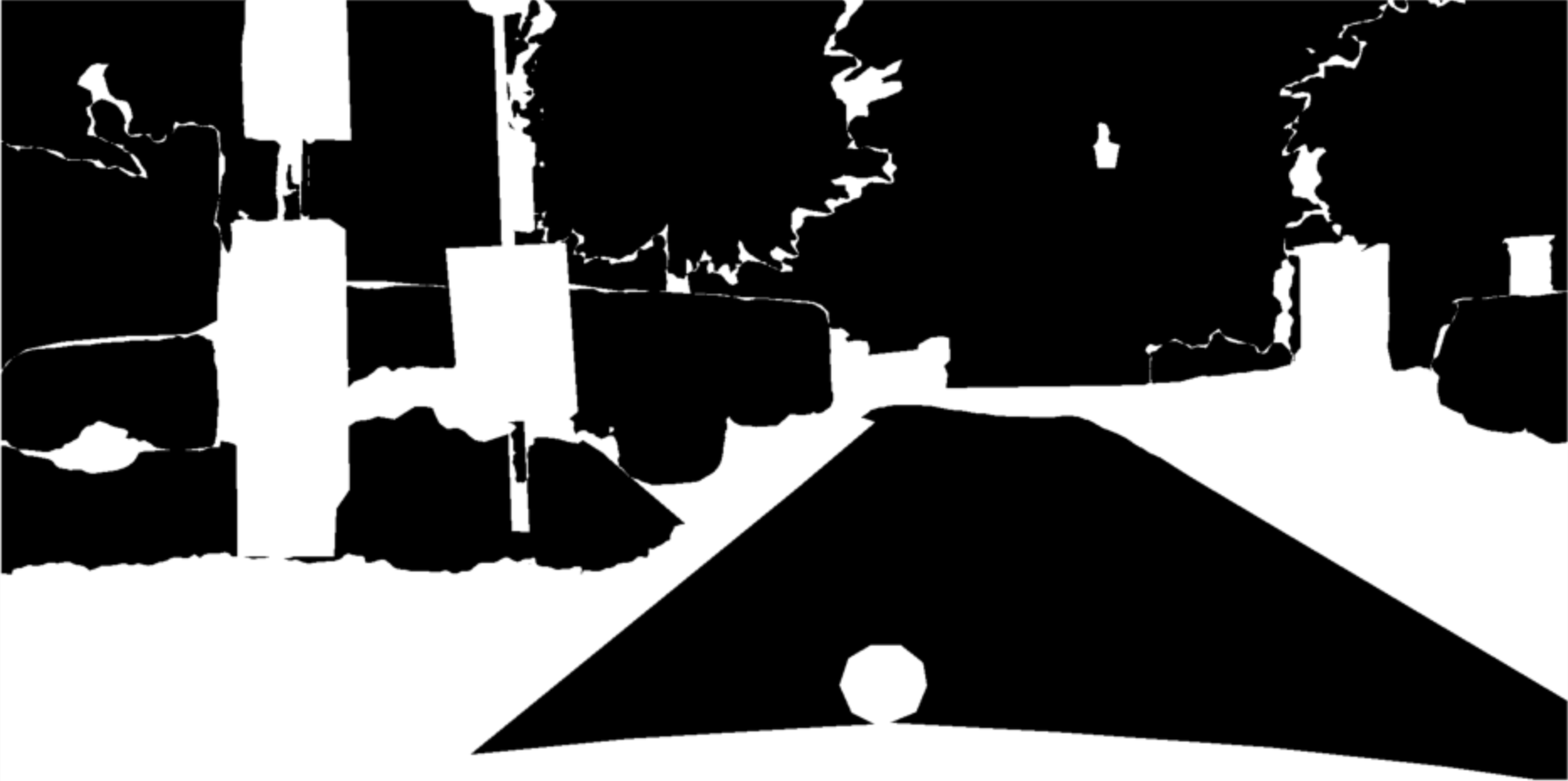}
        \caption{Binary Accuracy Map}
    \end{subfigure}
    \begin{subfigure}{0.195\textwidth}
        \includegraphics[width=\textwidth]{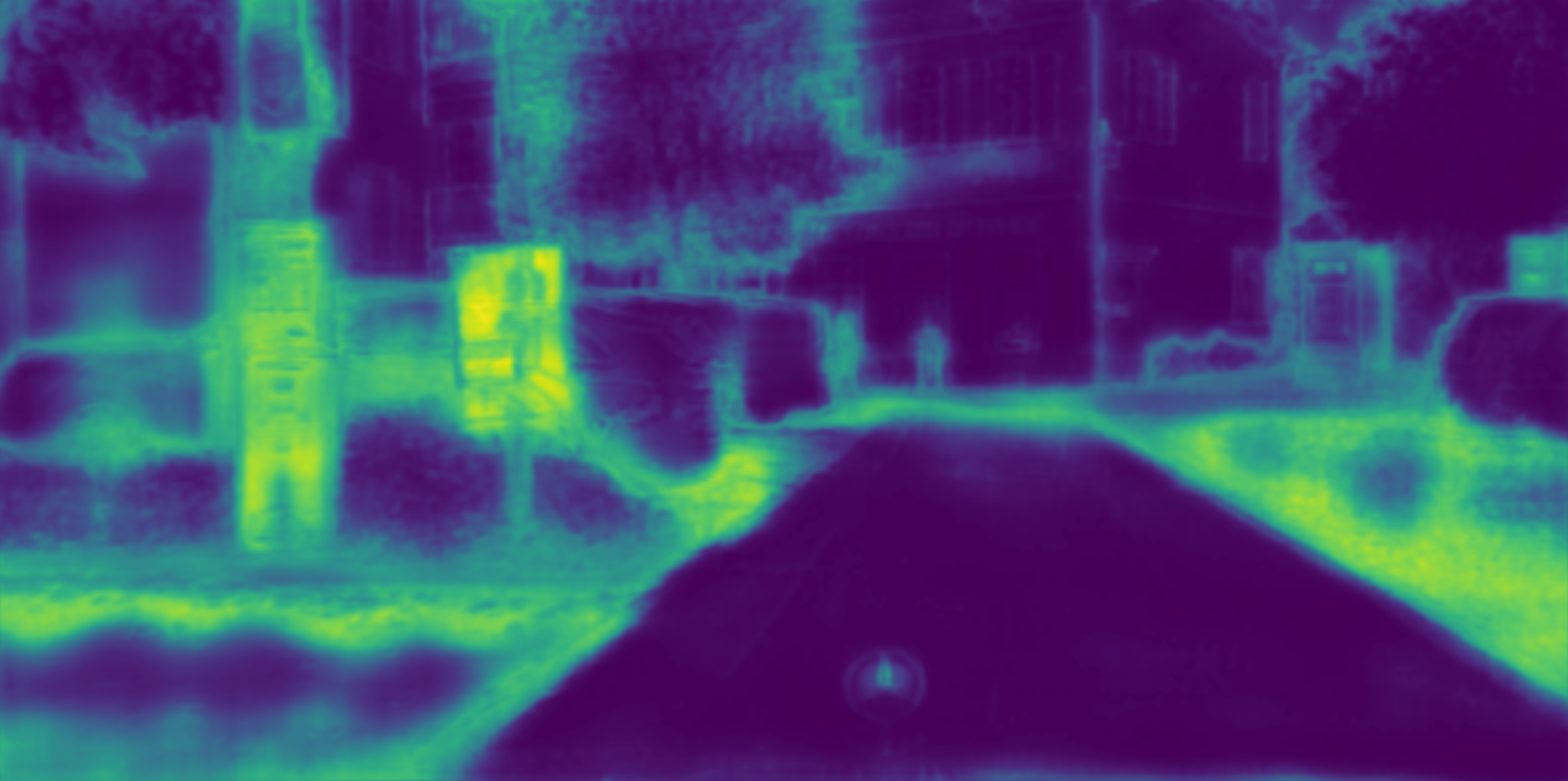}
        \caption{Uncertainty Prediction}
    \end{subfigure}
\caption{Example images from the Cityscapes validation set (a) with corresponding ground truth labels (b), our student's segmentation predictions (c), a binary accuracy map (d), and the student's uncertainty prediction (e). White pixels in the binary accuracy map are either incorrect predictions or void classes, which appear black in the ground truth label. For the uncertainty prediction, brighter pixels represent higher predictive uncertainties.}
\label{fig: qualitative_results}
\end{figure*}

Table \ref{table: class_wise_iou} and Table \ref{table: class_wise_uncertainty} outline a quantitative comparison between the student's and the teacher's Intersection over Union (IoU) as well as their predictive uncertainties. The results of Holder and Shafique \cite{holder2021EfficientUncertainty} have been included as they are the most relevant previous work on \ac{de}-based student-teacher distillation for efficient \ac{uq}. Their teacher is based on 25 DeepLabv3+ models with a MobileNet backbone \cite{howard2017MobileNetsEfficient}. The MobileNet backbone and our ResNet-18 backbone have been shown to have very similar performance \cite{bianco2018BenchmarkAnalysis}.

\textbf{Segmentation Prediction.} As shown in Table \ref{table: class_wise_iou}, our student network outperforms the teacher on the segmentation task for all classes except \textit{wall} and \textit{sky}, with an average improvement of 2.5\% in mIoU. We attribute this improvement to the student's ImageNet pre-training as compared to the randomly initialized ensemble members. The student by Holder and Shafique \cite{holder2021EfficientUncertainty} outperforms its teacher on the segmentation task for only one class and showed a mIoU deterioration of 4.2\%. 

\textbf{Uncertainty Prediction.} Table \ref{table: class_wise_uncertainty} shows that our student's approximation of the teacher's uncertainties is very accurate: In 10 out of the 19 classes our student's class-wise uncertainties deviate by less than 0.01 compared to the teacher's. Our student manages to deviate by less than 0.03 in 17 out of the 19 classes, with a maximum deviation of 0.042 for the \textit{bicycle} class. On the other hand, the student by Holder and Shafique \cite{holder2021EfficientUncertainty} deviates by less than 0.01 in 5 out of the 19 classes and by less than 0.03 in only 13 out of the 19 classes. Their student's maximum difference is 0.130 for the \textit{train} class. On average across all classes, both students' uncertainties deviate only slightly from those of the teachers, with our student model deviating by 0.002 and the student by Holder and Shafique \cite{holder2021EfficientUncertainty} deviating by -0.007. Generally speaking, both students struggle with accurately approximating the teacher's uncertainties for the last five classes: \textit{Truck}, \textit{bus}, \textit{train}, \textit{motorbike}, and \textit{bicycle}. For these classes, our student has an average absolute deviation of 0.028, while the student by Holder and Shafique \cite{holder2021EfficientUncertainty} deviates by 0.066.

\begin{figure}
    \centering
    \includegraphics[width=\linewidth]{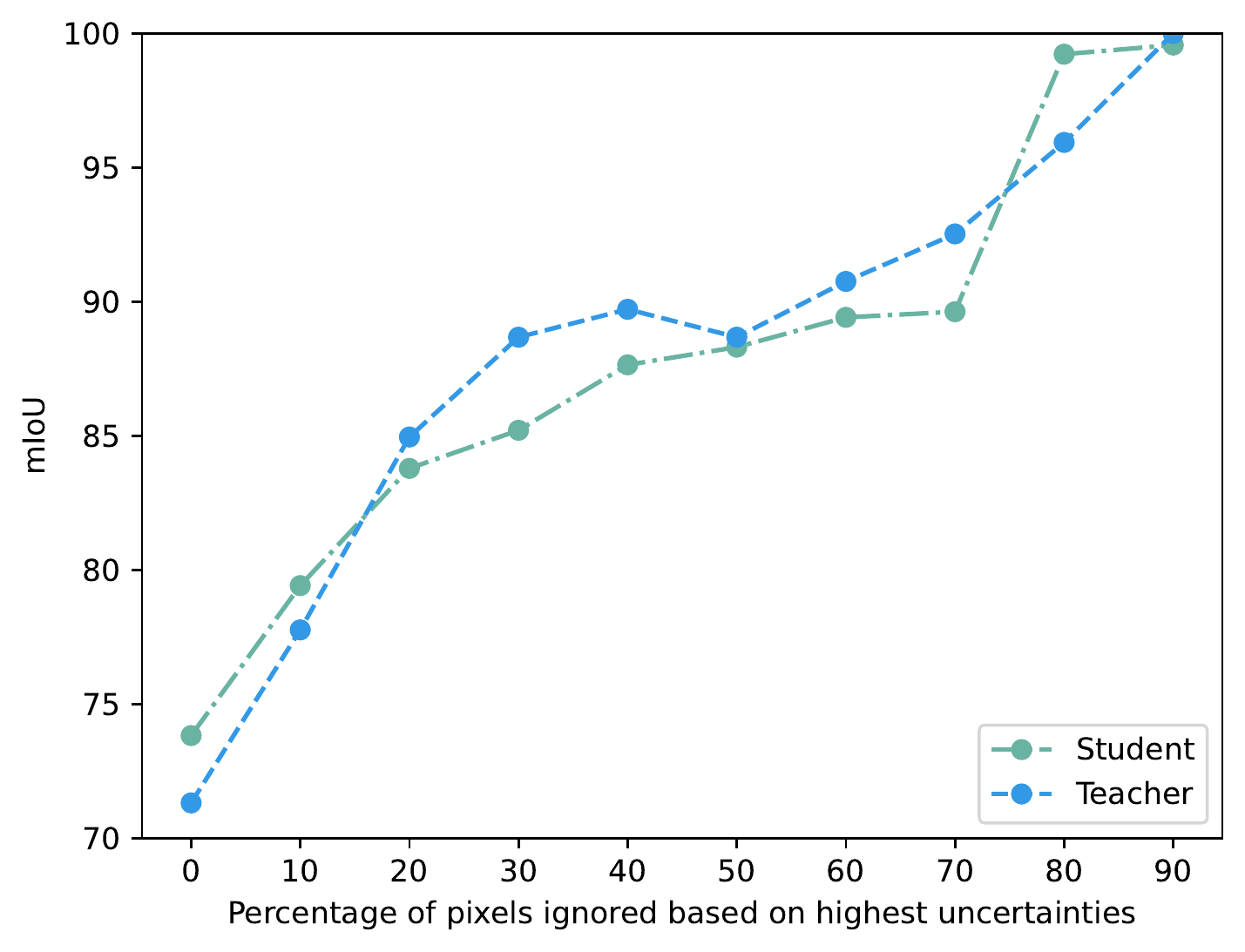}
    \caption{Comparison between the student's and the teacher's mean Intersection over Union (mIoU). We progressively ignore an increasing percentage of pixels in the segmentation prediction and simultaneously re-evaluated the mIoU. The pixels are sorted based on their predictive uncertainty in descending order, thus removing the most uncertain segmentation predictions first.}
    \label{fig: uncertaint_pixels_removed}
\end{figure}

Figure \ref{fig: uncertaint_pixels_removed} displays another comparison between the student's and the teacher's ability to approximate reliable uncertainties: For this analysis, we progressively ignored an increasing percentage of pixels in the segmentation prediction and simultaneously re-evaluated the mIoU. Thereby, the pixels were sorted based on their predictive uncertainty in descending order. This initally removes the pixels with the most uncertain segmentation predictions from the evaluation until only the pixels with the most certain predictions are left. Consequently, meaningful uncertainties should result in a monotonically increasing function.

As Figure \ref{fig: uncertaint_pixels_removed} shows, the student as well as the teacher experience an almost linear rise in mIoU from 73.8\% and 71.3\%, respectively, to almost 100\% after removing 90\% of the most uncertain pixels. Both models attain a similar relative increase in mIoU by disregarding the first 10\% of the most uncertain pixels. Up until ignoring 70\% of pixels, the teacher reaches a mIoU of 92.5\%, while the student only attains 89.6\%. Beyond this point, the student's mIoU surpasses the teacher's, with the student achieving 99.2\% after ignoring 80\% of the pixels with the highest uncertainties, while the teacher only reaches 95.9\%. This analysis yields two key findings: Firstly, predictive uncertainties prove to be an effective approach of identifying misclassified pixels. Secondly, our student's predictive uncertainties deviate only slightly from the teacher's uncertainties, revealing that they are equally meaningful.

\begin{table}[t!]
\begin{center}
\begin{tabular}{lrr}
& Inference time [ms] & Trainable Parameters\\ \hline
Baseline & $18.3 \pm 0.4$ & 12,333,923 \\
Teacher & $217.1 \pm 0.8$ & 123,339,230 \\
Student & $18.5 \pm 0.4$ & 12,334,180 \\
\end{tabular}
\end{center}
\caption{Comparison of the inference time for a single image in milliseconds and the number of trainable parameters between the baseline, the teacher, and the student model. The inference time and corresponding standard deviation are based on 25 independent runs.}
\label{table: inference}
\end{table}

\textbf{Inference Time.} Table \ref{table: inference} compares the inference time for a single image and the number of trainable parameters between the baseline, the teacher, and the student model. The experiment was conducted on a common NVIDIA GeForce RTX 3090 GPU with 24 GB of memory. What stands out the most is that there is only an insignificant difference of 0.2 milliseconds in inference time between the baseline and the student, despite the student's ability to output an additional predictive uncertainty. With 18.5 milliseconds per image, the student's inference time is roughly 11.7 times faster than the 217.1 milliseconds of the teacher. Table \ref{table: inference} also illustrates the number of trainable parameters, highlighting the efficiency of the student network. The additional uncertainty head of the student network only adds 257 parameters to the baseline model.

\subsection{Qualitative Evaluation}
Figure \ref{fig: qualitative_results} displays four example images from the Cityscapes validation set and their corresponding ground truth labels, our student's segmentation prediction, a binary accuracy map, and the student's uncertainty prediction. The binary accuracy map visualizes incorrectly predicted pixels and void classes in white and correctly predicted pixels in black.

Visually, for large areas and well-represented classes like road, sidewalk, building, sky, and car the student's performance on the segmentation task is almost free of errors. This supports the quantitative evaluation described in Table \ref{table: class_wise_uncertainty}. Like most segmentation models, our student struggles with class transitions, areas with lots of inherent noise, or areas that belong to the void class, which is visualized by the binary accuracy map.

A comparison of the binary accuracy map and our student's uncertainty prediction adds to the observations laid out in Table \ref{table: class_wise_uncertainty} and Figure \ref{fig: uncertaint_pixels_removed}: The uncertainty prediction reliably predicts high uncertainties for wrongly classified pixels and out-of-domain samples, which are visualized as white pixels in the binary accuracy map. For example, in the first image of Figure \ref{fig: qualitative_results}, our student correctly predicts high uncertainties in the noisy parts of the background and for fine geometric structures like traffic lights. Conversely, the student predicts very low uncertainties for the road, buildings, sky and vegetation. The second image example confirms this observation and adds two valuable insights about the quality of the student's uncertainty predictions. Firstly, although the train in the left part of the image is predicted correctly for the most part, the student still predicts high uncertainties. This is intuitively comprehensible and desired because the train class is underrepresented in the dataset and therefore potentially more difficult to detect reliably. Secondly, the student predicts high uncertainties in the bottom part of the image where reflections on the hood of the car cause incoherent segmentation predictions. The third image exemplifies another quality of the student's predictive uncertainty. In this case, the student struggles to correctly segment the truck in the right part of the image. Simultaneously, the student predicts high uncertainties for the entire truck, thus highlighting the flawed segmentation prediction. The fourth image demonstrates the student's capability to correctly identify out-of-domain samples, with high uncertainties predicted in areas belonging to the void class.

\subsection{Ablation Studies}
\begin{figure}[t!]
    \centering
    \includegraphics[width=\linewidth]{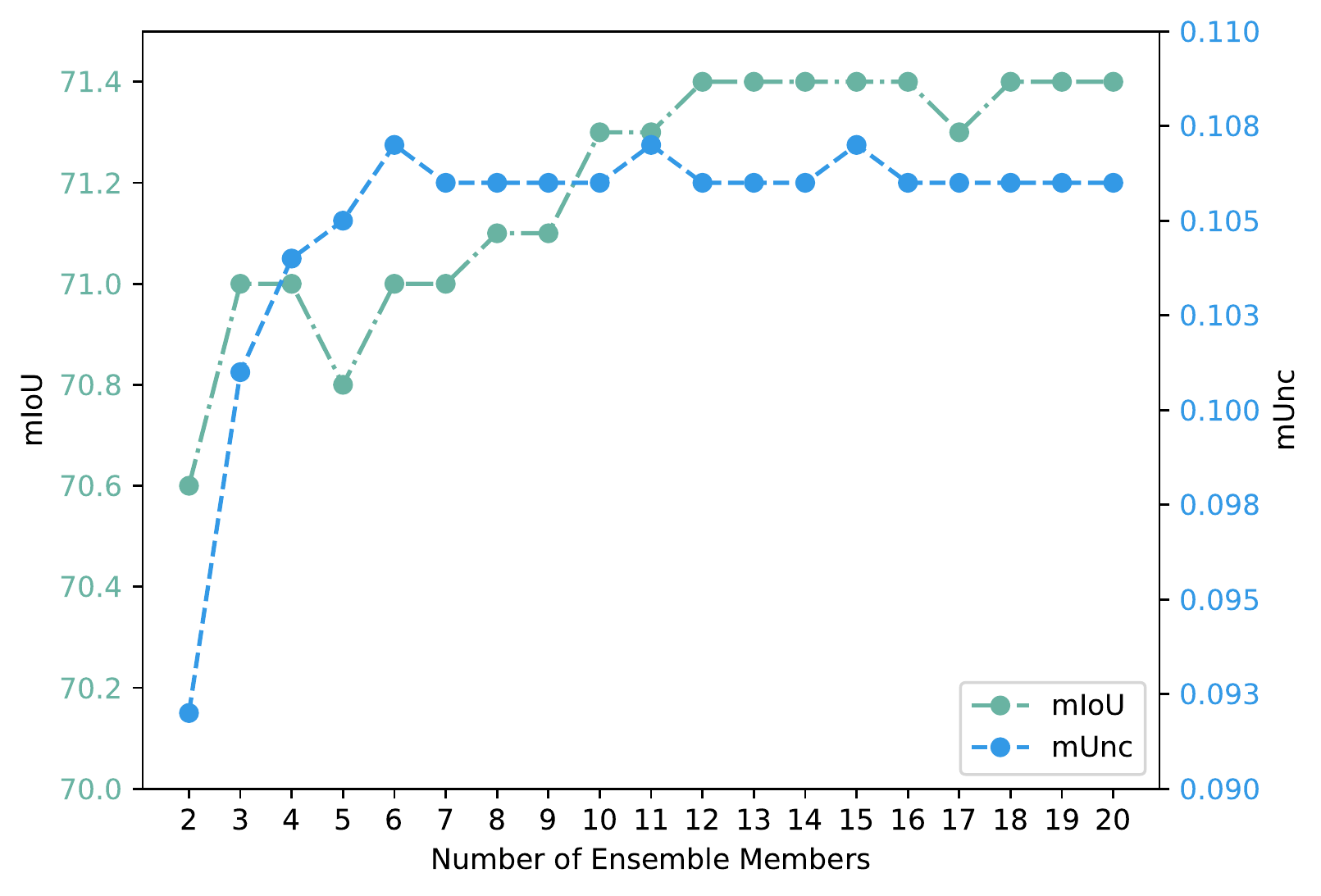}
    \caption{Ablation study on the impact of the number of ensemble members on the mean Intersection over Union (mIoU) and mean Uncertainty (mUnc).}
    \label{fig: ablation_study_ensemble}
\end{figure}

\begin{table}[t!]
\begin{center}
\begin{adjustbox}{width=\linewidth}
\begin{tabular}{lrrr}
& Training Epochs & mIoU $\uparrow$ & mUnc \\ \hline
Teacher$_{Random, n=10}$ & 200 & 71.3 & 0.106 \\ \hdashline
Baseline$_{Random}$ & 200 & 68.5 & - \\
Student$_{Random}$ & 200 & 64.6 & 0.097 \\
Baseline$_{ImageNet}$ & 200 & 73.7 & - \\
Student$_{ImageNet}$ & 200 & 73.8 & 0.108 \\ \hdashline
Student$_{Random}$ & 800 & 73.9 & 0.105 \\
\end{tabular}
\end{adjustbox}
\end{center}
\caption{Ablation study on the impact of ImageNet \cite{deng2009ImageNetLargescale} pre-training on the mean Intersection over Union (mIoU) and mean Uncertainty (mUnc).}
\label{table: ablation_study_iou}
\end{table}

\textbf{Number of Ensemble Members.} An essential part of \ac{dudes} is the quality of the teacher's uncertainty prediction which are based on a diverse set of ensemble members. Figure \ref{fig: ablation_study_ensemble} demonstrates the impact of the number of ensemble members on the mIoU and mean Uncertainty (mUnc). Naturally, adding more ensemble members improves the segmentation results. The mIoU increases from 70.6\% when using just two ensemble members to a maximum 71.4\% for twelve members. More importantly for \ac{dudes}, the mUnc increases from 0.092 for just two ensemble members to a maximum of 0.107 for six members. Adding more ensemble members to the teacher does not change the uncertainty prediction significantly as the mUnc stays within 0.106 and 0.107 until all twenty members are included. These findings go along with prior work on \ac{de}-based uncertainty quantification by Fort \etal and Lakshminarayanan \etal \cite{fort2020DeepEnsembles, lakshminarayanan2017SimpleScalable}. Consequently, we propose to use ten ensemble members for \ac{dudes}. 

\textbf{ImageNet Pre-training.} Table \ref{table: ablation_study_iou} shows the results of another ablation study on the impact of ImageNet \cite{deng2009ImageNetLargescale} pre-training on the mIoU and mUnc. We comprehensively compare the baseline segmentation model with our student model and our teacher, which consists of ten randomly initialized baseline models. The study does not examine the impact of ImageNet pre-training on the ensemble members as this would lead to less reliable uncertainties compared to random initialization \cite{fort2020DeepEnsembles, lakshminarayanan2017SimpleScalable}. 

While training for 200 epochs and using random initialization, our student underperforms the baseline model by 3.9\% and the teacher by 6.7\% with a mIoU of 64.6\% on the segmentation task. Our randomly initialized student also underestimates the teacher's uncertainties by 0.009 with a mUnc of 0.097. When using ImageNet pre-training for the baseline model and our student, both significantly improve their mIoU with 73.7\% and 73.8\% respectively. The student also manages to approximate the predictive uncertainties better with a mUnc of 0.108, which is close to the 0.106 of the teacher. It is worth noting that similar performance can also be achieved by randomly initializing our student when the number of training epochs is quadrupled to 800. This concurs with the findings of He \etal\cite{He_2019_ICCV}. As a consequence, we suggest using ImageNet pre-training for the student to improve convergence speed. 

\section{DISCUSSION}\label{sec: discussion}
\ac{dudes} applies student-teacher distillation with a \ac{de} to accurately approximate predictive uncertainties with a single forward pass while maintaining simplicity and adaptability. Against the teacher, the needed inference time per image is reduced by an order of magnitude and the computational overhead in comparison to the baseline is neglectable. Additionally, the student showed impressive capabilities of identifying wrongly classified pixels and out-of-domain samples within an image which is crucial for safety-critical applications such as autonomous driving and many other computer vision tasks.

In contrast to the work by Holder and Shafique \cite{holder2021EfficientUncertainty}, \ac{dudes} requires no major changes to the student's architecture and introduces only a single uncertainty loss without additional hyperparameters, yet delivers significant improvements over their work. Firstly, our student model outperforms its teacher in the segmentation task while their student suffers from a segmentation performance degradation in comparison to its teacher. Secondly, our student approximates its teacher's predictive uncertainties more closely than the student model by Holder and Shafique \cite{holder2021EfficientUncertainty}. More precisely, their student tends to underestimate uncertainties for classes with high uncertainties and vice versa, whereas our student does not suffer from any systematic shortcomings.  

A major factor of the effectiveness of \ac{dudes} lies in the simplification of what is distilled. Instead of distilling the teacher's uncertainty map, which is what Holder and Shafique \cite{holder2021EfficientUncertainty} proposed, we only use the teacher's predictive uncertainty. The teacher's uncertainty map is calculated by computing the standard deviation of the teacher's model's softmax probability maps along the class dimension. In the case of multiclass semantic segmentation, the resulting uncertainty map has dimensions of $C \times H \times W$, where $C$ is the number of classes, $H$ is the image height, and $W$ is the image width. For \ac{dudes}, the class dimension is reduced to $1$ by only considering the uncertainty of the predicted class in the segmentation map. Due to this simplification, the student's segmentation performance is not hindered and the predictive uncertainties can be learned more accurately. 

We acknowledge the simplification in the uncertainty distillation to be a potential limitation of \ac{dudes} as the student is only capable of estimating the uncertainty of the predicted class. However, there are practically no negative implications of this limitation since the remaining uncertainties are usually discarded anyway. Hence, \ac{dudes} remains useful for efficiently estimating predictive uncertainties for a wide range of applications while being easy to adapt.  

We believe that \ac{dudes} has the potential to provide a new promising paradigm in reliable uncertainty quantification by focusing on simplicity and efficiency. Except for the computational overhead during training, we found no apparent reason to not employ our proposed method in semantic segmentation applications where safety and reliability are critical.

\section{CONCLUSION}\label{sec: conclusion}
In this work, we propose \ac{dudes}, an efficient and reliable uncertainty quantification method by applying student-teacher distillation that maintains simplicity and adaptability throughout the entire framework. We quantitatively demonstrated that \ac{dudes} accurately captures predictive uncertainties without sacrificing performance on the segmentation task. Additionally, qualitative results indicate impressive capabilities of identifying wrongly classified pixels and out-of-domain samples.  With DUDES, we managed to simultaneously simplify and
outperform previous work on \ac{de}-based \ac{uq}.

We hope that \ac{dudes} encourages other researchers to incorporate uncertainties into state-of-the-art semantic segmentation approaches and to explore the usefulness of our proposed method for other tasks such as detection or depth estimation.

\section*{Acknowledgment}
The authors acknowledge support by the state of Baden-Württemberg through bwHPC.

This work is supported by the Helmholtz Association Initiative and Networking Fund on the HAICORE@KIT partition.

{\small
\bibliographystyle{ieee_fullname}
\bibliography{egbib}
}

\end{document}